%% file: neurips_2026.tex
\documentclass{article}

\PassOptionsToPackage{numbers, compress}{natbib}
\usepackage{wrapfig}
\usepackage{float}

\newif\ifarxiv
\arxivtrue
\ifarxiv
  \usepackage[preprint]{neurips_2026}
\else
  \usepackage[main,final]{neurips_2026}
\fi

\usepackage[utf8]{inputenc} % allow utf-8 input
\usepackage[T1]{fontenc}    % use 8-bit T1 fonts
\usepackage{hyperref}       % hyperlinks
\usepackage{url}            % simple URL typesetting
\usepackage{booktabs}       % professional-quality tables
\usepackage{amsfonts}       % blackboard math symbols
\usepackage{nicefrac}       % compact symbols for 1/2, etc.
\usepackage{microtype}      % microtypography
\usepackage[table]{xcolor}         % colors

\usepackage{graphicx}

\newcommand{\papertablestyle}{%
  \small
  \setlength{\tabcolsep}{4pt}%
  \renewcommand{\arraystretch}{1.15}%
}
\usepackage{subfigure}

\usepackage{enumitem}
\usepackage{algorithm}
\usepackage{algpseudocode}
\usepackage{colortbl}
\usepackage{pgf}
\hypersetup{pdftitle={RuleSmith: Multi-Agent LLMs for Automated Game Balancing},
  pdfauthor={Ziyao Zeng and coauthors},colorlinks=true,linkcolor=black,citecolor=black,urlcolor=blue}
\ifarxiv
\makeatletter
\renewcommand{\@noticestring}{40th Conference on Neural Information Processing Systems (NeurIPS 2026).}
\makeatother
\fi

\usepackage{listings}
\usepackage{amsmath}
\usepackage{amssymb}
\usepackage{mathtools}
\usepackage{amsthm}

\usepackage[capitalize,noabbrev]{cleveref}

\theoremstyle{plain}
\newtheorem{theorem}{Theorem}[section]
\newtheorem{proposition}[theorem]{Proposition}
\newtheorem{lemma}[theorem]{Lemma}

\theoremstyle{definition}

\theoremstyle{remark}

\newcommand{\cc}[1]{%
  \pgfmathsetmacro{\x}{min(max(#1,0),100)}%
  \pgfmathsetmacro{\t}{abs(\x-50)/50}%
  \ifdim \t pt < 0.5pt
    \pgfmathsetmacro{\u}{2*\t}%
    \pgfmathsetmacro{\r}{(205 + (240-205)*\u)/255}%
    \pgfmathsetmacro{\g}{(235 + (240-235)*\u)/255}%
    \pgfmathsetmacro{\b}{(205 + (240-205)*\u)/255}%
  \else
    \pgfmathsetmacro{\u}{2*(\t-0.5)}%
    \pgfmathsetmacro{\r}{(240 + (245-240)*\u)/255}%
    \pgfmathsetmacro{\g}{(240 + (210-240)*\u)/255}%
    \pgfmathsetmacro{\b}{(240 + (210-240)*\u)/255}%
  \fi
  \edef\cellcol{\noexpand\cellcolor[rgb]{\r,\g,\b}}%
  \cellcol #1%
}

\title{\fontsize{17.28}{20}\selectfont RuleSmith: Multi-Agent LLMs for \\ Automated Game Balancing}

\author{%
  \bfseries Ziyao Zeng$^{1}$ \quad Hao Wang$^{2}$ \quad Chen Liu$^{1}$ \quad Youheng Yao$^{1}$ \quad Jingcheng Ni$^{3}$ \\
  \bfseries Tianyu Liu$^{1}$ \quad Xiatao Sun$^{1}$ \quad Fengyu Yang$^{1}$ \quad Chenyu You$^{4}$ \quad Xiaofeng Liu$^{1}$ \\
  \bfseries Daniel Rakita$^{1}$ \quad Ronald R. Coifman$^{1}$ \quad Yuval Kluger$^{1}$ \quad Zhiwen Fan$^{2}$ \\[5pt]
  $^{1}$Yale University \quad $^{2}$Texas A\&M University \quad
  $^{3}$Brown University \quad $^{4}$Stony Brook University \\[2pt]
  \small\texttt{ziyao.zeng@yale.edu}
}

\begin{document}

% Keep paragraph and equation spacing natural on pages with floats.
\raggedbottom
\maketitle

\begin{abstract}
Game balancing is a longstanding challenge requiring repeated playtesting, expert intuition, and extensive manual tuning. We introduce \emph{RuleSmith}, a framework that couples multi-agent LLM self-play with surrogate-based optimization over a parameterized rule space. As a proof of concept, we build \emph{CivMini}, a simplified civilization-style game with two asymmetric factions governed by 12 tunable parameters. LLM agents interpret textual rulebooks and game states to propose actions that are validated by the game engine, yielding noisy estimates of faction win rates. Our primary implementation uses Gaussian-process Bayesian optimization with Expected-Improvement-based adaptive sampling to allocate more evaluation games to promising candidates. RuleSmith consistently finds near-balanced configurations: three independent optimization runs reach near-even empirical win rates in independent 200- and 300-game evaluations, and additional experiments with NTBEA and TPE show that the same self-play interface supports different search components. Cross-model evaluations and small human studies provide preliminary evidence of transfer across player populations. RuleSmith produces interpretable numerical rule adjustments without task-specific policy training. Project and code: \url{https://adonis-galaxy.github.io/RuleSmith-website/}.
\end{abstract}

%%%%%%%%%%%%%%%%%%%%%%%%%%%%%%%%
% INTRODUCTION
%%%%%%%%%%%%%%%%%%%%%%%%%%%%%%%%
\section{Introduction}

Balancing asymmetric strategy games is a core challenge in game design and multi-agent research. Factions differ in abilities, mobility, and economic structure, making it hard to tell whether a given ruleset produces fair or strategically rich gameplay. Traditional approaches rely on human experts cycling through manual tuning, heuristic adjustments, and subjective playtesting. These workflows are slow, expensive, and hard to scale. As games grow more complex, this process becomes not just time-consuming, but often intractable.

Many applications beyond games, including economic modeling, policy analysis, and cybersecurity, face a similar challenge: outcomes depend on parameterized rules, and assessing how rule changes propagate through multi-agent interactions is expensive. We believe the core idea of LLM self-play as a rule evaluator could extend to such settings, though our current work is a proof of concept in a controlled game environment; broader applications remain future work.

Large language models (LLMs) offer a promising new approach. LLMs can interpret rulebooks, reason over game states, and act as zero-shot simulators without task-specific policy training. Prior work shows that LLM agents can negotiate, plan over multiple steps, and exhibit complex strategic behavior~\cite{xu2023language,xu2023exploring,park2023generative}. LLM-assisted game design also uses language models to propose parameter edits from simulation feedback~\citep{huang2026game}. We study a complementary role: LLM agents supply the gameplay evaluations, while a separate optimizer searches the rules.

We introduce \textbf{RuleSmith}, a framework for automatically balancing asymmetric games through LLM self-play and Bayesian optimization. As a proof of concept, we built \textbf{CivMini}, a fully parameterized, grid-based, turn-based game inspired by civilization-style mechanics. Two asymmetric factions, Empire and Nomads, have distinct units, production systems, and combat stats. RuleSmith exposes tunable parameters governing HP, damage, resource gathering, production costs, and scoring weights. This shifts the object of optimization from agent policies to the game rules themselves.

To evaluate a candidate ruleset, RuleSmith runs two LLM agents and collects win rates from multiple self-play games. Since LLM gameplay evaluation is expensive and noisy, we pair Bayesian optimization with acquisition-based adaptive sampling: the number of games scales with each candidate's Expected Improvement, concentrating compute where balance improvement is most expected.

RuleSmith discovers configurations with near-even empirical faction win rates from imbalanced starting settings. Three independent optimization runs and larger held-out evaluations quantify the repeatability and uncertainty of these results. The optimized parameters describe how HP, resource efficiency, and production costs jointly affect outcomes under the chosen player population.

\begin{figure}[t]
    \centering
    \begin{minipage}[t]{0.54\linewidth}
        \vspace{0pt}
        \includegraphics[width=\linewidth]{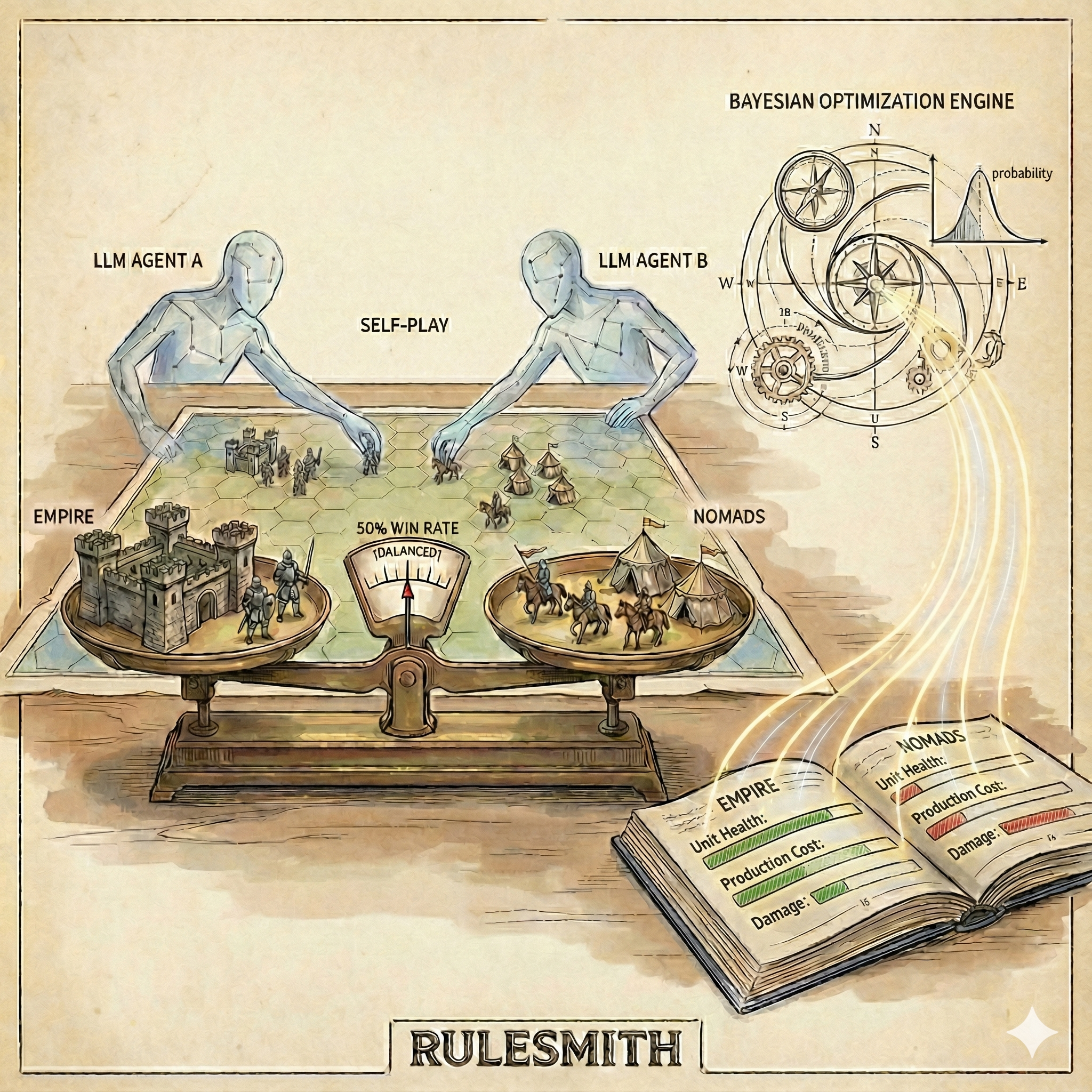}
    \end{minipage}%
    \hfill
    \begin{minipage}[t]{0.43\linewidth}
        \vspace{0pt}
        \hyphenpenalty=10000
        \exhyphenpenalty=10000
        \caption{%
            \textbf{Overview of RuleSmith.}
            RuleSmith treats asymmetric game balancing as optimization over game rollouts using multi-agent LLMs given a natural-language rulebook. It optimizes interpretable parameters such as unit HP, attack damage, resource gain, production cost, and scoring weights. For each candidate game parameter setting, two LLM agents play the game zero-shot, and their outcomes estimate the faction win rates and balance loss. A Bayesian optimizer then proposes new settings, while adaptive sampling allocates more games to promising candidates. This enables closed-loop balancing without task-specific policy training, reducing manual playtesting and parameter tuning.
            Figure generated by Nano Banana Pro.%
        }
        \label{fig:rulesmith_overview}
    \end{minipage}
    \vspace{-5mm}
\end{figure}

\paragraph{Contributions.}

\begin{itemize}[topsep=0pt, itemsep=4pt, parsep=0pt]
    \item \textbf{Zero-shot self-play from natural-language rulebooks.}
    LLM agents play CivMini from a rulebook and structured game states without task-specific training, with legality enforced by the engine.

    \item \textbf{Multi-agent pipeline for automated game balancing.}
    RuleSmith makes balance queryable through LLM self-play and connects this evaluator to interchangeable search components. Our primary implementation uses Gaussian-process Bayesian optimization and Expected-Improvement-based adaptive evaluation budgets.

    \item \textbf{Empirical validation across models and game configurations.}
    RuleSmith consistently reaches near-balance: three independent runs yield near-even empirical win rates, with confidence intervals from larger re-evaluations. We study stronger optimizer baselines, model and map variations, action validity, trajectory diversity, and transfer to small human-player samples.

\end{itemize}

%%%%%%%%%%%%%%%%%%%%%%%%%%%%%%%%
% RELATED WORK
%%%%%%%%%%%%%%%%%%%%%%%%%%%%%%%%
\section{Related Work}
\label{sec:related}

\paragraph{Game design automation.}
AI-based playtesting evaluates and refines board, trading-card, and video games using scripted or learned players~\citep{de2017ai,zhao2020winning,bergdahl2020augmenting}. Bayesian optimization has been used for game testing~\citep{celemin2024bayesian}; multi-fidelity methods motivate allocating different evaluation budgets to expensive candidates~\citep{kandasamy2016gaussian}. Procedural content generation (PCG) creates levels, maps, and mechanics~\citep{shaker2016procedural,hendrikx2013procedural,togelius2013procedural}; PCGML learns from examples~\citep{summerville2018procedural}, while discriminative and player-aware approaches guide content evaluation~\citep{karth2019addressing,snodgrass2017player,snodgrass2017studying}. RuleSmith searches numerical rules of an asymmetric executable game, with LLM self-play supplying the evaluations.

\paragraph{LLMs as designers versus evaluators.}
Huang et al.~\citep{huang2026game} present AI Game Designer, which uses LLM-guided reasoning to adjust level parameters from the statistical outputs of simulation. In that architecture, the LLM proposes design edits. RuleSmith instead uses LLMs as competing players: their rollouts provide a noisy balance objective, and an external optimizer proposes rule configurations. The engine executes and validates actions, while the LLM players supply behavior conditioned on the rulebook. These roles are complementary. Agent-based game-parameter tuning also predates LLMs: NTBEA was developed for noisy, expensive game evaluation~\citep{kunanusont2017ntbea,lucas2018ntbea}. Our contribution is the empirical side of this question: whether such a loop converges to balance at all, measured with multi-seed runs, optimizer baselines, large-sample re-evaluations, and significance tests. To our knowledge, this assignment of roles has not been studied for asymmetric rule spaces.

\paragraph{Multi-agent self-play.}
Self-play reinforcement learning primarily optimizes policies under fixed rules. AlphaGo and AlphaZero master board games~\citep{silver2016mastering,silver2017mastering,silver2018general}, while OpenAI Five and AlphaStar address complex multiplayer environments~\citep{berner2019dota,vinyals2019grandmaster}. CFR, neural fictitious self-play, and PSRO study strategic play and equilibrium approximation~\citep{brown2019deep,brown2019superhuman,heinrich2016deep,lanctot2017unified}; opponent-learning and population-based approaches model competitive adaptation~\citep{foerster2017learning,jaderberg2019human}. We optimize the rules rather than train player policies, using instruction-following LLMs as the evaluation population.

\paragraph{LLM agents.}
LLM-agent surveys cover tool use, autonomous planning, and simulation~\citep{wang2024survey,xi2025rise,gao2024large}. ReAct and HuggingGPT coordinate reasoning and actions~\citep{yao2022react,shen2023hugginggpt}, while CAMEL, generative agents, and Mindstorms organize interacting agents~\citep{li2023camel,park2023generative,zhuge2023mindstorms}. ALYMPICS studies strategic interaction in an LLM game-theory playground~\citep{mao2023alympics}, and language models can themselves act as optimizers~\citep{yang2023large} or compete in debate~\citep{liang2024encouraging,arnesen2024training}. RuleSmith grounds interaction in engine-validated game actions and connects the resulting outcomes to rule optimization.

%%%%%%%%%%%%%%%%%%%%%%%%%%%%%%%%
% METHOD
%%%%%%%%%%%%%%%%%%%%%%%%%%%%%%%%
\begin{figure}[t]
    \centering
    \includegraphics[width=\linewidth]{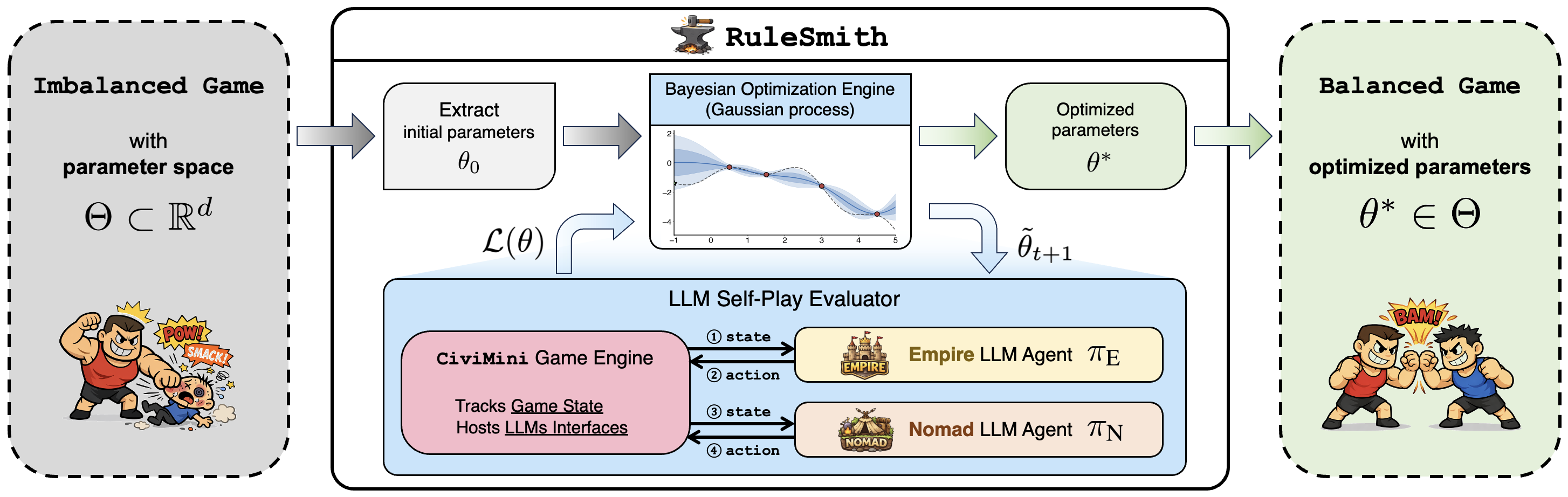}
    \vspace{-16pt}
    \caption{
        \textbf{Overview of the RuleSmith method.}
        We represent an asymmetric, turn-based strategy game (CivMini) as a parameterized rule space $\theta \in \Theta$, covering economy, combat, production, and scoring parameters.
        Given a candidate rule configuration $\theta_t$, two role-specific LLM agents (\emph{Empire} and \emph{Nomads}) play $N_t$ self-play games in the CivMini environment, producing a balance loss $\mathcal{L}(\theta)
=
|w_E(\theta)-0.5|+|w_N(\theta)-0.5|
+0.5\,w_D(\theta)$, where $w_E$, $w_N$, and $w_D$ are the Empire win rate, Nomads win rate, and draw rate, respectively.
        A Bayesian optimizer maintains a surrogate model $g(\theta)$ over a continuous relaxation of the rule space and selects new candidates $\tilde{\theta}_{t+1}$ by maximizing an acquisition function.
        The number of games $N_t$ is adaptively determined based on the Expected Improvement: promising candidates receive more games for accurate evaluation.
        Each continuous proposal is mapped to a valid, discrete ruleset via a deterministic discretization operator $D(\cdot)$ before evaluation. Cartoons in this figure were generated using ChatGPT-5.2.
    }
    \label{fig:method}
    \vspace{-3mm}
\end{figure}

\section{Method}
\label{sec:method}

RuleSmith takes five inputs: (i) an engine that executes actions and returns outcomes, (ii) a parameterized rule space with bounds and native granularities, (iii) a textual rulebook, (iv) an instruction-following LLM that proposes actions, and (v) a scalar balance objective. The engine executes the rules and validates actions; the LLM agents supply player decisions. The optimizer queries the same self-play interface regardless of its search algorithm. CivMini instantiates this interface in a controlled setting.

We consider balancing an asymmetric, parameterized turn-based strategy game so that two factions have approximately equal win probabilities under a specified LLM-player population. Let $\theta \in \Theta \subset \mathbb{R}^d$ denote a vector of game parameters (e.g., unit damage, production cost, score weights), and let $\pi_{\text{E}}$ and $\pi_{\text{N}}$ denote the policies of the LLM agents playing \emph{Empire} and \emph{Nomads}.

For a fixed parameterization $\theta$, let $w_E(\theta)$, $w_N(\theta)$, and $w_D(\theta)$ denote the probabilities that a rollout ends in an Empire win, a Nomads win, or a draw, respectively. These probabilities satisfy
\begin{equation}
    w_E(\theta) + w_N(\theta) + w_D(\theta) = 1.
\end{equation}
In practice, we estimate them from $N$ self-play games. For example, the empirical Empire win rate is
\begin{equation}
    \widehat{w}_{E}(\theta)
    =
    \frac{1}{N} \sum_{i=1}^{N} \mathbb{1}\{ \text{Empire wins game } i \mid \theta \}.
\end{equation}
The quantities $\widehat{w}_{N}(\theta)$ and $\widehat{w}_{D}(\theta)$ are estimated analogously.

Our goal is to choose $\theta$ so that faction win probabilities are close to 0.5 under the specified player population, while avoiding frequent draws. We therefore use the balance loss
\begin{equation}
    \mathcal{L}(\theta)
    =
    |w_E(\theta)-0.5|+|w_N(\theta)-0.5|
    +0.5\,w_D(\theta).
    \label{eq:balance-objective}
\end{equation}
The first two terms measure deviations from the target win probabilities. With no draws, their sum equals $|w_E-w_N|$. When draws occur, $w_E+w_N<1$, so draws also affect the target-deviation terms; the explicit $0.5w_D$ term further discourages unresolved games. This term is a safeguard that keeps the optimizer away from configurations in which neither side can win; the coefficient is a design choice, not a tuned optimum, and because the term is almost always zero, any moderate value would serve the same purpose. Draws are rare in the additional evaluations (0/500 in one and 2/300 in another), so this term contributes at most a few thousandths to the measured loss. City destruction or score-based resolution determines almost all outcomes.

In practice, we treat $\mathcal{L}$ as an expensive and noisy black-box function: each evaluation of $\mathcal{L}(\theta)$ requires multiple LLM-driven game simulations and is affected by stochasticity in model decisions and game dynamics. We therefore optimize~\eqref{eq:balance-objective} with Bayesian optimization over a continuous relaxation of $\Theta$, combined with a discrete projection that maps suggested parameters to valid game configurations.

\subsection{Parametric Asymmetric Game: CivMini}
\label{subsec:civmini}

To study asymmetric game balancing in a controlled setting, we introduce \emph{CivMini}, a minimalistic two-faction, turn-based asymmetric strategy game inspired by 4X games\footnote{4X games: A subgenre of strategy games characterized by eXplore, eXpand, eXploit, and eXterminate, emphasizing large-scale empire management and long-horizon planning.} such as \emph{Civilization}~\cite{civ}. CivMini is implemented as a deterministic, grid-based environment with a fully observable state and a small but expressive set of unit types and actions.

\paragraph{Map, factions, and units.}
The game is played on a $7\times 7$ grid with two asymmetric factions: \emph{Empire} and \emph{Nomads}. Each starts with a city (Empire at $(1,1)$, Nomads at $(5,5)$) and units at fixed coordinates. Empire specializes its mobile units: \texttt{Farmers} gather resources and \texttt{Soldiers} fight. Nomads deploy a single combat unit type, \texttt{Cavalry}, with higher mobility (2 cells per turn versus 1 for Empire units). Cavalry cannot \texttt{GATHER}, but Nomads earn resources from kills, incentivizing aggression. Cities of both factions can also actively \texttt{PRODUCE\_RESOURCE}. Farmers have fixed HP (5.0); Soldier and Cavalry HP are optimizable.

\paragraph{Actions and turn structure.}
Play proceeds in discrete turns, up to a fixed maximum number of turns
$T_{\max}$. In the main experiments, $T_{\max}$ is fixed and is not part of
the 12-parameter rule space; Appendix~\ref{subsec:supp_map_size} studies game-design variants with different map sizes and corresponding turn limits. In each turn, players alternate controlling all of their units. We use a multi-action mode in which \emph{each unit} of the current player performs exactly one action per turn, allowing a single model response to coordinate the units while the engine resolves their actions sequentially. For each unit, the action space includes:
\begin{itemize}[topsep=0pt, itemsep=4pt, parsep=0pt]
    \item \texttt{GATHER}: collect resources from the current tile (Farmers only),
    \item \texttt{MOVE}: move up to the unit's movement allowance in Manhattan distance to an unoccupied tile (Cavalry move up to 2 cells, Farmers and Soldiers up to 1; Cities cannot move),
    \item \texttt{BATTLE}: attack an adjacent enemy unit or city, dealing parameterized damage (Soldiers, Cavalry, and Cities can battle),
    \item \texttt{PRODUCE\_RESOURCE}: cities generate resources,
    \item \texttt{PRODUCE\_UNIT}: cities spend resources to spawn new units in adjacent empty cells,
    \item \texttt{PASS}: take no action.
\end{itemize}
One LLM call proposes the current player's actions jointly, reducing inference cost relative to calling the model for every unit. Actions then resolve in a fixed order (city first, followed by units in creation order), and each is rechecked against the updated state. Thus, a city cannot spend resources gathered later in the same player turn. An earlier move can also invalidate a later move, in which case the later action becomes \texttt{PASS}. A single-action mode is available, but all main experiments use joint action proposals.

The engine enforces hard constraints such as one action per unit per turn, one unit per tile (no stacking), adjacency requirements for combat, and immobile city tiles. City tiles are modeled as attackable entities with hit points, making it possible to win by conquest.

\vspace{-4pt}
\paragraph{Victory and scoring.}
Games may end early if one faction destroys the opposing city. Otherwise, when the turn limit is reached, the game is resolved by a score-based tie-breaker. The final score for each faction is a weighted combination of (i) remaining resources, (ii) the number of battles won, and (iii) the number of surviving units, each controlled by a corresponding parameter in $\theta$:
\begin{equation}
\begin{aligned}
\text{score} =\;
& \theta_{\text{res}} \cdot \text{resources}
+ \theta_{\text{battle}} \cdot \text{battles\_won} \\
& + \theta_{\text{unit}} \cdot \text{surviving\_units}.
\end{aligned}
\end{equation}
The balance score is defined in Eq.~\eqref{eq:balance-objective}; it penalizes deviations from the target win rate and draws.

\vspace{-4pt}
\paragraph{Parameterization.}
CivMini exposes $d=12$ tunable parameters that jointly define the game rules:
\begin{itemize}[topsep=0pt, itemsep=4pt, parsep=0pt]
    \item \emph{Economy} (3 parameters): initial resources, Empire farmer gather amount, and Nomads kill resource gain;
    \item \emph{Combat} (4 parameters): damage per attack and HP for Empire Soldiers and Nomad Cavalry;
    \item \emph{Production} (2 parameters): unit production cost in resources for each faction;
    \item \emph{Scoring} (3 parameters): weights for resources, battles won, and surviving units.
\end{itemize}
Table~\ref{tab:parameters} gives the parameter ranges and grids. Mobility, initial units, and Farmer HP remain fixed to preserve faction identity; setting Farmer HP to 5 maintains vulnerability to raids. The remaining numerical strength, economy, cost, and score parameters are searched. This partition and its bounds are designer inputs, and optima at the boundaries can motivate expanding the ranges.

\subsection{LLM Self-Play as an Evaluator}
\label{subsec:llm_selfplay}

Given a parameterized game instance $\theta$, we use two LLM agents to play the Empire and Nomads roles and estimate the resulting balance. At each decision point, the environment constructs a structured natural-language description of the game state, including the current turn index, each faction's resources and unit positions, a faction-specific strategy guide, and the list of legal actions. This description is concatenated with a role-specific system prompt and rules retrieved from a lightweight RAG system (TF-IDF + cosine similarity). Since we use multi-action mode, the agent is asked to produce a JSON object containing actions for \emph{all} units simultaneously:
\begin{quote}
    \texttt{\{"empire\_farmer\_0": \{"action\_type": "GATHER"\},}\\
    \texttt{\ "empire\_city": \{"action\_type": "PRODUCE\_UNIT",}\\
    \texttt{\ \ "produce\_unit\_type": "soldier", "to": \{"x": 1, "y": 2\}\}\}}
\end{quote}
The engine parses and validates the response: a unit whose action is missing, unparseable, or outside its enumerated legal set receives a fallback legal action chosen according to a fixed priority (\texttt{BATTLE}, \texttt{GATHER}, \texttt{MOVE}, \texttt{PASS}) with random tie-breaking. Actions are checked again during sequential execution, and any action invalidated by an earlier action in the same turn becomes \texttt{PASS}. The opponent acts after the current player finishes.

Strategy guides supply state-dependent heuristic action recommendations, and JSON examples specify the output format. Zero-shot refers to the absence of task-specific policy training (Appendix~\ref{sec:supp_reproducibility}).
For each candidate $\theta$, we run $N$ games with independent rollout randomness. Stochastic decoding (temperature 0.7, top-$p$ 0.9) varies the actions and thus the trajectories under identical parameters. Appendix~\ref{sec:supp_agent_diagnostics} quantifies the resulting diversity and execution-level conflicts. Engine validity does not imply optimal action selection or the absence of strategically biased play.

\subsection{Bayesian Optimization over Rule Space}
\label{subsec:bo}

Enumerating the roughly $1.9\times10^{10}$ discrete configurations would require prohibitively many LLM rollouts. We therefore employ Bayesian optimization~\cite{snoek2012practical} over a continuous relaxation of the rule space.

We maintain a Gaussian process (GP) surrogate model $g(\theta)$ of the balance loss $\mathcal{L}(\theta)$, implemented with scikit-optimize~\citep{head2018skopt}. At iteration $t$, the optimizer selects a new candidate $\tilde{\theta}_t$ by maximizing an acquisition function over the history $\mathcal{D}_{t-1} = \{ (\theta_i, f(\theta_i)) \}_{i=1}^{t-1}$. Before evaluation, a deterministic projection $D$ maps $\tilde{\theta}_t$ to a valid discrete configuration $\theta_t = D(\tilde{\theta}_t)$ by rounding and clamping each coordinate to the native grid in Table~\ref{tab:parameters}: integer steps except for the resource-score weight, which uses steps of $0.1$. The surrogate is updated with the evaluated, discretized coordinates.

\vspace{-4pt}
\paragraph{Acquisition-based adaptive sampling.}
To improve efficiency, inspired by multi-fidelity optimization~\cite{swersky2013multi, kandasamy2017multi}, we introduce an adaptive sampling strategy that allocates more evaluation budget to promising candidates. At each iteration, we dynamically adjust the number of self-play games $N_t$ based on the Expected Improvement (EI) of the proposed candidate:
\begin{equation}
    \text{EI}(\theta) = (y^* - \mu(\theta)) \Phi(z) + \sigma(\theta) \phi(z),
\end{equation}
where $y^*$ is the best observed value, $\mu(\theta)$ and $\sigma(\theta)$ are the GP posterior mean and standard deviation, $z = (y^* - \mu(\theta)) / \sigma(\theta)$, and $\Phi$ and $\phi$ are the standard normal CDF and PDF. The selected candidate receives an integer number of games based on its EI relative to the running maximum. Let $e_t=\operatorname{EI}(\tilde{\theta}_t)$, $M_t=\max_{i\leq t} e_i$, and $\epsilon>0$ prevent division by zero. Then
\begin{equation}
 N_t=\operatorname{clip}_{[N_{\min},N_{\max}]}\!\left(
 \left\lfloor N_{\min}+(N_{\max}-N_{\min})
 \min\!\left\{\frac{e_t}{\max(M_t,\epsilon)},1\right\}\right\rfloor\right).
 \label{eq:adaptive-games}
\end{equation}
We use $N_{\min}=16$ and $N_{\max}=64$; the released implementation uses $\epsilon=10^{-3}$. The first 10 random initialization points receive $N_{\min}$ games because no fitted posterior is available. Zero EI also yields $N_{\min}$. EI is computed on the pre-evaluation posterior, so $N_t$ is fixed before observing the candidate's outcomes. This is an allocation heuristic whose practical benefit is assessed empirically, rather than an optimal allocation guarantee.

\vspace{-6pt}
\section{Results}
\label{sec:experiments}

We first summarize what the theory does and does not cover (Section~\ref{subsec:theoretical_results}) and provide the implementation details (Section~\ref{subsec:implementation}). We then report the main balancing results across agent pairings (Section~\ref{subsec:main_results}), their repeatability and statistical uncertainty (Section~\ref{subsec:statistical_results}), ablations of the search component and the player population (Section~\ref{sec:ablation}), and extended game configurations (Section~\ref{sec:generalization}).

\vspace{-6pt}
\subsection{Theoretical Results}
\label{subsec:theoretical_results}

\vspace{-6pt}
Our theory concerns balance estimation and idealized GP-based search; it does not guarantee the implemented procedure. Proposition~\ref{prop:symmetry} shows that exact balance is attainable under role symmetry when the draw probability is zero, while Lemma~\ref{lemma:concentration_of_loss} shows that the empirical projected balance loss from self-play concentrates around the true projected loss at rate $O(N_t^{-1/2})$. Proposition~\ref{prop:idealized_gp_lcb} provides an idealized GP-LCB regret bound under explicit calibration and information-gain assumptions. It is a statement about the balance loss and the GP model class: under these assumptions, some GP-based rule has vanishing average regret. It is not a guarantee for EI or for adaptive replication, which we evaluate empirically, including an LCB variant (Section~\ref{subsec:statistical_results}). Statements and proofs are in Appendix~\ref{sec:supp_theory}.

\vspace{-6pt}
\subsection{Implementation Details}
\label{subsec:implementation}

\paragraph{Model and infrastructure.}
We use InternVL3.5~\cite{internvl3.5} in FP16 as the decision-making agent for both factions. Many current model families are released as vision-language models, and InternVL3.5 provides 2B and 8B checkpoints with the same architecture, tokenizer, and prompt format, allowing agent capacity to be varied without changing anything else. Inputs and outputs are text only; the visual encoder is never invoked and no result depends on visual input, although the interface would permit inputs such as rendered game maps in future work. We combine the 2B and 8B checkpoints into the four faction pairings in Table~\ref{tab:ewin_nwin_matrix}. Game optimizations and evaluations are parallelized across 8 NVIDIA A100 80GB GPUs. Optimization using InternVL3.5-8B for both sides over 100 iterations takes around 40 hours. Several experiments were run on NVIDIA RTX 6000 Ada GPUs. For generalization experiments (Appendix~\ref{sec:supp_generalization}), we additionally use Qwen3.5-9B as an optimization agent and Gemini~2.5~Flash and Gemini~2.5~Pro as evaluation agents.

\begin{table}[t]
\papertablestyle
\caption{\textbf{Optimizable parameters in CivMini.} We optimize 12 parameters governing economy, combat, production, and scoring. Precision indicates the discretization granularity applied before evaluation.}
\label{tab:parameters}
\vspace{1mm}
\centering
\begin{tabular}{lcclcc}
\toprule
\textbf{Parameter} & \textbf{Range} & \textbf{Prec.} &
\textbf{Parameter} & \textbf{Range} & \textbf{Prec.} \\
\midrule
\multicolumn{3}{l}{\emph{Economy}} & \multicolumn{3}{l}{\emph{Production}} \\
Initial resources    & $[2, 10]$    & int & Empire unit cost  & $[2, 10]$    & int \\
Empire farmer gather & $[1, 5]$     & int & Nomads unit cost  & $[2, 10]$    & int \\
Nomads kill gain & $[1, 10]$    & int &                   &              &     \\
\midrule
\multicolumn{3}{l}{\emph{Combat}} & \multicolumn{3}{l}{\emph{Scoring}} \\
Empire damage        & $[1, 5]$     & int & Resource weight   & $[0.1, 0.5]$ & 0.1 \\
Nomads damage        & $[1, 5]$     & int & Battle weight     & $[1, 5]$     & int \\
Empire soldier HP    & $[4, 16]$    & int & Unit weight       & $[1, 5]$     & int \\
Nomads cavalry HP    & $[4, 16]$    & int &                   &              &     \\
\bottomrule
\end{tabular}
\vspace{-2mm}
\end{table}

\vspace{-6pt}
\paragraph{Optimization setup.}
We optimize 12 parameters (Table~\ref{tab:parameters}) with a Mat\'ern-$5/2$ GP and EI over 100 iterations, beginning with 10 random points; proposals are projected before evaluation and before the surrogate update. The remaining settings are conventional rather than tuned. Temperature 0.7 and top-$p$ 0.9 are standard sampling values under which play varies across games while staying coherent. A fixed query retrieves three rule entries, keeping the prompt short. The 16-turn cap only bounds worst-case cost, since most games end earlier (Appendix~\ref{sec:supp_agent_diagnostics}). Sixteen games per candidate correspond to two evaluation waves on 8 GPUs, and 64 caps the cost of a single candidate. Parameter bounds are coarse brackets (Table~\ref{tab:parameters}; for example, Soldier HP 4--16 around the released default of 8), not tuned intervals. Our budget did not permit ablating each of these choices; we prioritized the ablations most likely to change conclusions (model scale, map size, agent type, and the search component). Appendix~\ref{sec:supp_reproducibility} lists the settings in full.

\vspace{-6pt}
\paragraph{Evaluation protocol.}
During optimization, for each candidate $\theta$, we run $N_t \in [16, 64]$ independent self-play games. Games terminate when a city is destroyed (instant victory) or the turn limit is reached (score-based resolution). The balance loss $\mathcal{L}(\theta)$ is given in Eq.~\eqref{eq:balance-objective}. The main experiments evaluate selected checkpoints on 100 held-out games. The additional statistical study evaluates three independent optimization runs on separate 200- and 300-game batches and the optimized configuration of Table~\ref{tab:supp_human_vs_opt} on 500 games. These held-out evaluations are separate from the candidate evaluations used by the optimizer: in-loop evaluations are intentionally noisy, which the adaptive game count manages, and all reported numbers come from held-out evaluations. The $50\%\pm5\%$ criterion refers to observed win rates, not a confidence-certified population bound.

\begin{table}[!b]
\papertablestyle
\vspace{3mm}
\caption{\textbf{Optimized winning chances of Empire and Nomads.} Rows specify the agent pairing used to optimize rules; columns specify the evaluation pairing. Each cell reports \textit{Empire wins} $\mid$ \textit{Nomads wins} over 100 held-out games using the best optimization checkpoint (empirical loss $\leq0.1$). \textit{E}/\textit{N} denote the factions; subscripts denote model sizes. Greener cells indicate more even outcomes; empirical near-balance ($50\%\pm5\%$) is bolded.}
\label{tab:ewin_nwin_matrix}
\vspace{1mm}
\centering
\begin{tabular}{lcccc}
\toprule
\textbf{Optimize $\backslash$ Evaluate}
& \textbf{\textit{E}$_{2\mathrm{B}}$ vs \textit{N}$_{2\mathrm{B}}$}
& \textbf{\textit{E}$_{2\mathrm{B}}$ vs \textit{N}$_{8\mathrm{B}}$}
& \textbf{\textit{E}$_{8\mathrm{B}}$ vs \textit{N}$_{2\mathrm{B}}$}
& \textbf{\textit{E}$_{8\mathrm{B}}$ vs \textit{N}$_{8\mathrm{B}}$} \\
\midrule
\textbf{\textit{E}$_{2\mathrm{B}}$ vs \textit{N}$_{2\mathrm{B}}$}
& \textbf{\cc{48}$\mid$52} & \cc{32}$\mid$68 & \cc{73}$\mid$27 & \textbf{\cc{55}$\mid$45}  \\

\textbf{\textit{E}$_{2\mathrm{B}}$ vs \textit{N}$_{8\mathrm{B}}$}
& \cc{81}$\mid$19 & \textbf{\cc{47}$\mid$53}  & \cc{91}$\mid$9\phantom{0} & \cc{75}$\mid$25 \\

\textbf{\textit{E}$_{8\mathrm{B}}$ vs \textit{N}$_{2\mathrm{B}}$}
& \cc{37}$\mid$63 & \phantom{0}\cc{6}$\mid$94 & \textbf{\cc{52}$\mid$48} & \cc{29}$\mid$71 \\

\textbf{\textit{E}$_{8\mathrm{B}}$ vs \textit{N}$_{8\mathrm{B}}$}
& \textbf{\cc{53}$\mid$47}  & \cc{24}$\mid$76 & \cc{81}$\mid$19 & \textbf{\cc{51}$\mid$49}  \\

\bottomrule
\end{tabular}
\end{table}

\subsection{Main Empirical Results}
\label{subsec:main_results}
Table~\ref{tab:ewin_nwin_matrix} shows win rates across optimization and evaluation pairings. Diagonal entries meet the empirical near-balance criterion. Off-diagonal, upgrading one side shifts outcomes in its favor; the largest shift occurs when reversing the optimization asymmetry (optimizing with \textit{E}$_{8\mathrm{B}}$ vs \textit{N}$_{2\mathrm{B}}$ and evaluating with \textit{E}$_{2\mathrm{B}}$ vs \textit{N}$_{8\mathrm{B}}$). Both unoptimized configurations we evaluated are decisively imbalanced: the authors' initial configuration (Appendix~\ref{sec:supp_human_vs_opt}) yields an Empire win rate of 100\%, and the default configuration of the released code yields 11.3\% for Empire over 300 games (Table~\ref{tab:statistical_evaluation}). The optimized configuration in Table~\ref{tab:supp_human_vs_opt} yields 51\%$\mid$49\% over 100 games and 47.6\%$\mid$52.4\% over 500 games.

The tested configurations also show empirical transfer across LLM families. Table~\ref{tab:cross_llm} shows that configurations optimized with InternVL3.5-8B or Qwen3.5-9B remain near-balanced when evaluated by the other family (balance score $\leq 0.04$ across all four matchups). Further generalization results, including Gemini~2.5~Flash/Pro and human players, are in Appendix~\ref{sec:supp_generalization}.

\vspace{-6pt}
\subsection{Repeatability and Statistical Uncertainty}
\label{subsec:statistical_results}
Three independent GP-based optimization runs vary optimization initialization and self-play randomness. At 200 games per final configuration, the mean empirical loss is $0.0267$ (sample SD $0.0058$); at 300 games, it is $0.0333$ (SD $0.0200$). All six point estimates fall within the empirical near-balance criterion. Table~\ref{tab:statistical_evaluation} gives the 300-game evaluations and a 500-game re-evaluation of the original configuration, with Wilson 95\% intervals and two-sided exact binomial tests. No configuration from the GP (EI), NTBEA, or TPE runs shows a detectable departure from equal win probability at these sample sizes. This supports repeatability in the tested setting, although failing to reject $p_E=0.5$ is not an equivalence test establishing $p_E\in[0.45,0.55]$.

\input{tables/statistical_evaluation.tex}

\paragraph{Acquisition rule.} We also ran the loop once with the GP lower confidence bound (LCB, with scikit-optimize's default $\kappa=1.96$) in place of EI, leaving everything else unchanged. The run drives the in-loop loss to zero, but its selected configuration is not near-balanced on held-out games: Empire wins 57.5\% of 200 games (Wilson 95\% CI [50.6, 64.1]) and 57.7\% of 300 games ([52.0, 63.1]). With a single run and a configuration selected from small in-loop batches, this does not rank the two rules; we keep EI because its value also sets the adaptive game count (Section~\ref{subsec:bo}).

At 100 games near 50\%, a 95\% interval spans roughly $\pm10$ percentage points; the other ablations remain single-run studies. Full 200-/300-game results and statistical definitions are in Appendix~\ref{sec:supp_statistics}.

\paragraph{Qualitative behavior.}
Figure~\ref{fig:game_vis} illustrates a Nomads conquest through mobility and sustained city pressure; further rollouts show both factions winning through different tactics (Appendix~\ref{subsec:representative_rollout}).

\begin{table}[!htbp]
\papertablestyle
\caption{\textbf{Cross-LLM-family transfer.} Empirical win rates meet the near-balance criterion ($50\%\pm5\%$) when parameters optimized with one LLM family are evaluated by the other. All results use 100 evaluation games.}
\label{tab:cross_llm}
% \vspace{2mm}
\centering
\begin{tabular}{llccc}
\toprule
\textbf{Optimization LLM} & \textbf{Evaluation LLM} & \textbf{Empire\%} & \textbf{Nomads\%} & $\mathcal{L}$ \\
\midrule
InternVL3.5-8B & InternVL3.5-8B & 51 & 49 & 0.02 \\
InternVL3.5-8B & Qwen3.5-9B     & 50 & 50 & 0.00 \\
Qwen3.5-9B     & InternVL3.5-8B & 49 & 51 & 0.02 \\
Qwen3.5-9B     & Qwen3.5-9B     & 52 & 48 & 0.04 \\
\bottomrule
\end{tabular}
\vspace{-3mm}
\end{table}

\subsection{Ablation Studies}
\label{sec:ablation}
\textbf{Different optimization methods.}
Table~\ref{tab:ablation_optimization} compares adaptive BO with Random Search, a (1+1) Evolution Strategy (ES), and fixed-budget GP-BO. Random Search samples uniformly from the discrete rule space; ES perturbs a single incumbent with isotropic Gaussian mutations and accepts improvements. Both use $N=64$ games per candidate, while fixed GP-BO uses $N=16$, $32$, or $64$. All methods run for 100 iterations and receive a 100-game held-out evaluation with 8B agents on both sides.
Adaptive BO and fixed GP-BO with $N=64$ reach empirical near-balance, using about 3500 and 6400 optimization games, respectively. The other variants do not meet this criterion. These single-run comparisons suggest a benefit from adaptive replication in this setup, though they do not establish a general optimizer ranking.

\begin{table}[!htbp]
\papertablestyle

\caption{\textbf{Original search and replication ablation.} Each method runs for 100 iterations and is evaluated on 100 held-out games. Random Search and (1+1)-ES use 64 games per candidate; GP-BO uses either fixed replication or 16--64 games based on EI. These are single optimization runs.}
\label{tab:ablation_optimization}
\vspace{2mm}
\centering
\begin{tabular}{lccc}
\toprule
& \textbf{Random Search}
& \textbf{(1+1)-ES}
& \textbf{BO (adaptive)} \\
\midrule
\textbf{Win Rate}
& \cc{13}$\mid$87
& \cc{26}$\mid$74
& \textbf{\cc{51}$\mid$49} \\
\midrule
& \textbf{BO ($N=16$)}
& \textbf{BO ($N=32$)}
& \textbf{BO ($N=64$)} \\
\midrule
\textbf{Win Rate}
& \cc{34}$\mid$66
& \cc{61}$\mid$39
& \textbf{\cc{48}$\mid$52} \\
\bottomrule
\end{tabular}
\vspace{-2mm}
\end{table}

\textbf{Stronger search components.}
We additionally evaluate the N-Tuple Bandit Evolutionary Algorithm (NTBEA)~\citep{lucas2018ntbea} and the Tree-structured Parzen Estimator (TPE)~\citep{bergstra2011algorithms} within the same engine, rulebook, and LLM self-play interface. At 200/300 evaluation games, Empire win rates are 49.0\%/48.0\% for NTBEA and 51.5\%/54.3\% for TPE, compared to means of 50.7\%/51.2\% across the three GP runs, which are comparable within uncertainty (Tables~\ref{tab:statistical_evaluation} and~\ref{tab:all_reevaluations}). All three reach comparable balance with the engine and self-play objective unchanged, showing that RuleSmith works with different surrogate-based search components; because they converge to similar balance on CivMini, this testbed does not rank them. The fidelity mechanism behind BOHB and Hyperband corresponds here to running more self-play games per candidate, which adaptive sampling already provides.

\textbf{Proxy player quality.}
LLM agents outperform random legal-action agents and provide meaningful competition to scripted agents, although such outcomes do not by themselves establish human-like or optimal play. Execution-level conflicts account for 3.3\% of attempted unit actions for 8B and 4.6\% for 2B agents; the engine converts these to \texttt{PASS}. An additional 300-game analysis finds 229 distinct three-turn openings and a mean turn-5 board-occupancy Jaccard distance of 0.62, supporting trajectory diversity under stochastic decoding (Table~\ref{tab:trajectory_diversity}).

\begin{table}[!htbp]
\papertablestyle
\centering\caption{\textbf{Trajectory diversity} under identical rules (8B self-play, NTBEA configuration; Appendix~\ref{sec:supp_agent_diagnostics}). The 100-game column is a subset of the 300-game column, not an independent replication.}
\label{tab:trajectory_diversity}
\begin{tabular}{lrr}
\toprule
Metric & First 100 games & All 300 games \\
\midrule
Distinct three-turn opening sequences & 86/100 & 229/300 \\
Distinct first-turn action combinations & 6 & 11 \\
Action entropy, Empire / Nomads (bits) & 2.03 / 1.65 & 2.04 / 1.64 \\
Mean turn-5 Jaccard distance & 0.62 & 0.62 \\
Ended before the turn cap (a city destroyed) & 63\% & 62\% \\
Game length, mean (SD) in turns & 10.8 (4.4) & 11.1 (4.4) \\
\bottomrule
\end{tabular}
\end{table}

Three participants played 120 games against the agents, with randomized faction and opponent scale and blinded scale assignment. Humans won 55/60 against 2B and 48/60 against 8B. Yet Empire won 45.0\% and 43.3\% of these games, and 53.3\% in the separate 60-game human-versus-human study; none of the three departs detectably from an even split (all $p\geq0.37$). The configuration optimized by LLM self-play is consistent with balance in these populations; given the few, repeated participants, this provides preliminary evidence (Appendices~\ref{sec:supp_user_study} and~\ref{subsec:human_vs_agent}).

\subsection{Generalization and Scalability}
\label{sec:generalization}

\paragraph{Scalability and extended configurations (Appendix~\ref{sec:supp_extended}).}
In the tested settings, RuleSmith finds near-even empirical outcomes across map sizes from 5$\times$5 to 11$\times$11 (Table~\ref{tab:extended}; Appendix~\ref{subsec:supp_map_size}). It also scales to three factions (Empire, Nomads, Theocracy), balancing all three simultaneously (balance loss 0.013--0.033). Grid size works as a discrete optimization variable; results stay near-balanced.

\begin{table}[!htbp]
\papertablestyle
\setlength{\tabcolsep}{3.2pt}
\caption{\textbf{Extended configurations.} Empire$\mid$Nomads win rates (with Theocracy third for three factions) and balance loss for four map sizes with InternVL3.5-8B (turn limits in parentheses), three-faction balancing with four added Theocracy parameters ($\mathcal{L}_3$), and map size (5--9) as an additional optimizable parameter (Appendix~\ref{sec:supp_extended}). Map-size losses follow Eq.~\eqref{eq:balance-objective} from the reported win rates.}
\label{tab:extended}
\centering
\begin{tabular}{lcccc@{\hspace{9pt}}cc@{\hspace{9pt}}cc}
\toprule
& \multicolumn{4}{c}{\textbf{Map size} (turn limit)} & \multicolumn{2}{c}{\textbf{Three factions}} & \multicolumn{2}{c}{\textbf{Map size optimized}} \\
\cmidrule(lr){2-5}\cmidrule(lr){6-7}\cmidrule(lr){8-9}
& 5$\times$5 (16) & 7$\times$7 (16) & 9$\times$9 (32) & 11$\times$11 (32) & 8B & 2B & 8B & 2B \\
\midrule
Win rate (\%) & 53$\mid$47 & 51$\mid$49 & 48$\mid$52 & 51$\mid$49 & 32$\mid$33$\mid$35 & 34$\mid$33$\mid$33 & 53$\mid$47 & 51$\mid$49 \\
Balance loss & 0.06 & 0.02 & 0.04 & 0.02 & 0.033 & 0.013 & 0.06 & 0.02 \\
\bottomrule
\end{tabular}
\end{table}

\vspace{-6pt}
\paragraph{Parameter analysis (Appendix~\ref{sec:supp_analysis}).}
Optimized parameters vary with agent pairing. Symmetric pairings yield similar TTK values in both directions; asymmetric pairings show coordinated changes in combat, economy, cost, and scoring, consistent with compensating for capacity differences. Distinct pairings yield different configurations with similar empirical balance scores (Table~\ref{tab:supp_optimized_parameters}).

%%%%%%%%%%%%%%%%%%%%%%%%%%%%%%%%
% CONCLUSION
%%%%%%%%%%%%%%%%%%%%%%%%%%%%%%%%
\vspace{-6pt}
\section{Conclusion, Limitations, and Impact Statement}
\label{sec:conclusion}

RuleSmith automates asymmetric game balancing through LLM self-play and Bayesian optimization over interpretable rule parameters. EI-based adaptive replication focuses evaluation on promising candidates. Three independent GP runs and larger held-out evaluations strengthen the balance results, while NTBEA/TPE demonstrate compatibility with other search components. Cross-model and small human studies suggest transfer across the tested populations. Balance remains population-dependent, and since CivMini is a controlled proof of concept, scaling to larger rule spaces, partial observability, and population shifts remains open (Appendices~\ref{sec:supp_limitations} and~\ref{sec:supp_impact}).

% Acknowledgments sit after the 10 content pages and before the references, as the camera-ready rules allow
\clearpage
\begin{ack}
The authors declare no competing interests.
\end{ack}

\bibliography{RuleSmith}
\bibliographystyle{unsrtnat}

%%%%%%%%%%%%%%%%%%%%%%%%%%%%%%%%%%%%%%%%%%%%%%%%%%%%%%%%%%%%%%%%%%%%%%%%%%%%%%
% APPENDIX
%%%%%%%%%%%%%%%%%%%%%%%%%%%%%%%%%%%%%%%%%%%%%%%%%%%%%%%%%%%%%%%%%%%%%%%%%%%%%%
\clearpage
\appendix

%%%%%%%%%%%%%%%%%%%%%%%%%%%%%%%%%%%%%%%%%%%%%%%%%%%%%%%%%%%%
\section{Limitations}
\label{sec:supp_limitations}

\paragraph{LLM-agent-centric balance criteria.}
RuleSmith optimizes for balance as measured by LLM self-play win rates. Cross-LLM evaluations and the small human study provide encouraging evidence of transfer to the tested populations (Appendix~\ref{sec:supp_generalization}). LLM win rates of 54--78\% against rule-based opponents establish an intermediate opponent baseline while also showing sensitivity to player pairing (Appendix~\ref{sec:supp_vs_rule}). Within these studies, faction outcomes show no detectable imbalance: human players defeat the agents in most games, yet faction outcomes stay statistically even against both agent scales and between humans (Appendix~\ref{sec:supp_user_study}). Formal guarantees of transfer to human gameplay remain absent, and these studies are limited in scale. Systematic human evaluation and player modeling would strengthen the transfer evidence.

\paragraph{Action validity, strategy bias, and skill.}
Validation ensures executable games but not good strategies. The measured 3.3\%/4.6\% execution-rejection rates reflect within-turn conflicts rather than a complete measure of hallucination or parse failure. Both factions are played by the same model, so population-level biases, such as a general preference for aggression, affect both sides and largely cancel in the win-rate disparity; what remains is that balance is defined relative to the evaluated population. The released prompts include strategy guides, so the evaluation population is defined jointly by the model, decoding, and prompts. Humans substantially outperform these agents; how large the strength gap between optimization and evaluation players can become before the optimized rules lose balance remains an open question for future work.

\paragraph{Statistical and reproducibility scope.}
The three-seed experiment covers the main GP setting. The other ablations and the additional NTBEA, TPE, and LCB searches are single runs. Selecting the final configuration by its lowest in-loop loss can favor a lucky small-batch evaluation; the held-out re-evaluations guard against this, as the single LCB run illustrates. Binomial uncertainty and failure to reject equality do not establish equivalence within five percentage points. Human games share a few participants and should not be treated as 60 independent draws from a broad human population. Seeds and optimizer settings for the additional runs are listed in Table~\ref{tab:settings}.

\paragraph{Designer-specified rule space.}
The engine, the choice of tunable versus fixed rules, parameter bounds, and the objective require designer input. A balanced solution must be accessible within those bounds. Several reported settings lie at boundaries; widening ranges and re-optimizing is a practical extension, while automated range discovery remains future work.

\paragraph{Simplified game environment.}
CivMini is a deliberately simplified testbed with a fully observable state, small action spaces, and a compact parameter set. Real commercial strategy games involve thousands of interacting parameters, imperfect information, long-horizon dependencies, and emergent player behaviors that our current framework does not address. Extending RuleSmith to richer, partially observable, or multi-stage environments is a key direction for future work.

\paragraph{No formal optimality guarantees.}
Bayesian optimization with a Gaussian process surrogate provides no guarantee of finding the globally optimal balanced configuration. The surrogate model may misfit in high-curvature regions, and the discrete projection step introduces approximation errors. Robustness under distribution shifts (such as changes in the player population or game meta) is not explicitly modeled. Incorporating robustness objectives or adversarial evaluation into the optimization loop is a promising avenue for future research.

%%%%%%%%%%%%%%%%%%%%%%%%%%%%%%%%%%%%%%%%%%%%%%%%%%%%%%%%%%%%
\section{Broader Impact}
\label{sec:supp_impact}
This work advances automated design tools and multi-agent simulation by enabling LLM-based evaluation and optimization of rule-driven systems. By treating large language models as flexible evaluators rather than deployed decision-makers, RuleSmith offers a scalable and interpretable approach to analyzing complex, parameterized environments.

Potential applications of this framework extend beyond games to domains such as economic modeling, policy analysis, and resource allocation, where outcomes depend on structured rules and multi-actor interactions. At the same time, we recognize that automated simulation and optimization of competitive or adversarial systems may raise ethical considerations if misapplied. In particular, such tools could be misused to over-optimize asymmetric rules in ways that disadvantage certain participants or reduce transparency in decision-making processes.

To mitigate these risks, RuleSmith is designed as an offline analysis and design-time tool operating on abstract, simplified environments, rather than a system for real-time decision execution. We emphasize that any real-world deployment of rule-optimization frameworks should incorporate human oversight, domain-specific constraints, and ethical review. Overall, we believe the primary impact of this work lies in supporting safer, more transparent, and more systematic design of rule-based systems, while encouraging responsible use in sensitive application domains.

%%%%%%%%%%%%%%%%%%%%%%%%%%%%%%%%%%%%%%%%%%%%%%%%%%%%%%%%%%%%
\section{Theoretical Analysis}
\label{sec:supp_theory}

This section provides an idealized analysis of two assumptions behind RuleSmith: when exact balance is attainable, and why Gaussian-process Bayesian optimization (GP-BO) is a suitable search procedure for the balance loss. These results should be read as theoretical context rather than as a finite-budget optimality guarantee for the full RuleSmith implementation, which uses Expected Improvement, adaptive evaluation budgets, and a discrete projection step.

\begin{proposition}[Balanced Optimum under Role Symmetry]
\label{prop:symmetry}
Consider a two-faction game parameterized by $\theta \in \Theta$. Suppose $\Theta$ contains a role-exchange-invariant configuration $\theta_{\mathrm{sym}}$, and suppose the proxy-player policy distribution is invariant under swapping the two faction roles. Then
\begin{equation}
    w_E(\theta_{\mathrm{sym}})
    =
    w_N(\theta_{\mathrm{sym}}).
\end{equation}
Consequently, for the balance loss in Eq.~\eqref{eq:balance-objective},
\begin{equation}
    \mathcal{L}(\theta_{\mathrm{sym}})
    =
    1.5\,w_D(\theta_{\mathrm{sym}}).
\end{equation}
In particular, if $w_D(\theta_{\mathrm{sym}})=0$, then
\begin{equation}
    \mathcal{L}(\theta_{\mathrm{sym}})=0,
\end{equation}
and the optimal balance loss obeys
\begin{equation}
    \min_{\theta \in \Theta} \mathcal{L}(\theta)=0.
\end{equation}
\end{proposition}

\begin{proof}
At a role-exchange-invariant configuration, swapping the faction labels leaves the game distribution unchanged. By the policy-symmetry assumption, the induced distribution over trajectories is also unchanged after the swap, except that Empire and Nomads exchange labels. Therefore, the probability of an Empire win before the swap equals the probability of a Nomads win after the swap, giving
\[
    w_E(\theta_{\mathrm{sym}})
    =
    w_N(\theta_{\mathrm{sym}}).
\]
Since $w_E=w_N=(1-w_D)/2$ at this configuration, substituting into Eq.~\eqref{eq:balance-objective} yields
\[
    \mathcal{L}(\theta_{\mathrm{sym}})
    =
    |w_E(\theta_{\mathrm{sym}})-0.5|+|w_N(\theta_{\mathrm{sym}})-0.5|
    +0.5\,w_D(\theta_{\mathrm{sym}})
    =
    1.5\,w_D(\theta_{\mathrm{sym}}).
\]
Thus, if the symmetric configuration has no draw probability, $w_D(\theta_{\mathrm{sym}})=0$, then $\mathcal{L}(\theta_{\mathrm{sym}})=0$. Since the balance loss is nonnegative by definition, this implies $\min_{\theta \in \Theta}\mathcal{L}(\theta)=0$.
\end{proof}

\emph{Remark.}
This proposition does not claim that every asymmetric ruleset can be made perfectly balanced. Role symmetry guarantees equal faction win probabilities, while exact zero loss for the draw-penalized objective additionally requires zero draw probability. RuleSmith targets the harder practical case where the searchable parameter subspace is asymmetric, discrete, and evaluated through noisy LLM self-play.
\begin{lemma}[Concentration of the empirical projected balance loss]
\label{lemma:concentration_of_loss}
    Let $\widetilde{\Theta}$ denote the continuous relaxation used by the optimizer,
let $D:\widetilde{\Theta}\rightarrow \Theta_{\mathrm{disc}}$ be the deterministic discretization operator, and define the projected balance objective
\[
    F(\tilde{\theta}) := \mathcal{L}(D(\tilde{\theta})).
\]
At round $t$, let $\tilde{\theta}_t \in \widetilde{\Theta}$ be the continuous proposal selected by the optimizer, and let
\[
    \theta_t = D(\tilde{\theta}_t) \in \Theta_{\mathrm{disc}}
\]
be the corresponding discretized rule configuration evaluated by the game engine. Suppose $N_t$ is chosen using only information available before observing
the round-$t$ self-play outcomes. From $N_t$ independent rollouts under
$\theta_t$, define
\[
    \hat{w}_{E,t}, \hat{w}_{N,t}, \hat{w}_{D,t}
\]
as the empirical Empire-win, Nomads-win, and draw rates, respectively, and let
\[
    y_t =
    |\hat{w}_{E,t}-0.5|+|\hat{w}_{N,t}-0.5| + 0.5\hat{w}_{D,t}
\]
be the empirical balance loss observed by the optimizer. Then, for any fixed
horizon $T$ and failure probability $\delta \in (0,1)$, with probability at least $1-\delta$, simultaneously for all $t=1,\ldots,T$,
\[
    |y_t - F(\tilde{\theta}_t)|
    =
    |y_t - \mathcal{L}(D(\tilde{\theta}_t))|
    \leq
    \eta_t,
    \qquad
    \eta_t =
    \frac{5}{2}
    \sqrt{
        \frac{\log(6T/\delta)}{2N_t}
    }.
\]
\end{lemma}
\begin{proof}
    Condition on the history before round $t$. Since $\tilde{\theta}_t$ and $N_t$ are chosen before observing the round-$t$ outcomes, the projected ruleset $\theta_t = D(\tilde{\theta}_t)$ is also fixed before these outcomes are observed. Therefore, the $N_t$ rollouts are independent samples from the outcome distribution induced by the evaluated ruleset $\theta_t$.

For each outcome type $j\in\{E,N,D\}$, Hoeffding's inequality gives
\[
    |\hat{w}_{j,t} - w_j(\theta_t)|
    \leq
    \sqrt{
        \frac{\log(6T/\delta)}{2N_t}
    }
\]
simultaneously for all $j\in\{E,N,D\}$ and all $t=1,\ldots,T$ with probability
at least $1-\delta$, by a union bound.

On this event, the inequality $||a-c|-|b-c||\leq |a-b|$ gives
\[
\begin{aligned}
 |y_t-F(\tilde{\theta}_t)|
 &\leq |\hat w_{E,t}-w_E(\theta_t)|
       +|\hat w_{N,t}-w_N(\theta_t)|
       +0.5|\hat w_{D,t}-w_D(\theta_t)| \\
 &\leq \frac{5}{2}\sqrt{\frac{\log(6T/\delta)}{2N_t}}.
\end{aligned}
\]
This proves the claim.
\end{proof}

\begin{proposition}[Idealized GP-LCB for projected self-play evaluations]
\label{prop:idealized_gp_lcb}
Let $\widetilde{\Theta}$ denote the continuous relaxation used by the optimizer, let $D:\widetilde{\Theta}\rightarrow \Theta_{\mathrm{disc}}$ be the deterministic discretization operator, and define the projected objective
\[
    F(\tilde{\theta}) := \mathcal{L}(D(\tilde{\theta})).
\]
Let
\[
    \tilde{\theta}^{\star}
    \in
    \arg\min_{\tilde{\theta}\in\widetilde{\Theta}} F(\tilde{\theta})
\]
be an optimal continuous representative of a best projected ruleset.

At round $t$, an idealized optimizer selects a continuous proposal $\tilde{\theta}_t\in\widetilde{\Theta}$, evaluates the projected ruleset $D(\tilde{\theta}_t)$ using $N_t$ independent self-play rollouts, and observes the empirical projected loss $y_t$ defined in Lemma~\ref{lemma:concentration_of_loss}. For this idealized projected-objective analysis, the surrogate is indexed by the pair $(\tilde{\theta}_t,y_t)$; the implementation stores evaluated discrete coordinates instead. This distinction is part of why this proposition is not a guarantee for the implemented pipeline.

Assume a bounded kernel diagonal $k(x,x)\leq1$, nondecreasing $\beta_t$, and a GP observation model with variances bounded above and bounded away from zero (the finite replication range permits such bounds). Assume that the GP surrogate is calibrated for the projected objective $F$ in the standard GP bandit sense: there exist confidence parameters
$\{\beta_t\}_{t\geq 1}$ such that, with probability at least $1-\delta$,
\[
    |F(\tilde{\theta})-\mu_{t-1}(\tilde{\theta})|
    \leq
    \sqrt{\beta_t}\sigma_{t-1}(\tilde{\theta})
\]
for all $\tilde{\theta}\in\widetilde{\Theta}$ and all $t\geq 1$. Here
$\mu_{t-1}$ and $\sigma_{t-1}$ are the GP posterior mean and posterior standard deviation after $t-1$ observations. Lemma~\ref{lemma:concentration_of_loss} motivates a heteroscedastic observation-uncertainty proxy scaling as $\eta_t^2=O(1/N_t)$, although the GP posterior remains an idealized modeling approximation.

Consider the idealized GP lower-confidence-bound rule
\[
    \tilde{\theta}_t
    \in
    \arg\min_{\tilde{\theta}\in\widetilde{\Theta}}
    \left[
        \mu_{t-1}(\tilde{\theta})
        -
        \sqrt{\beta_t}\sigma_{t-1}(\tilde{\theta})
    \right].
\]
Then, on the confidence event above, the cumulative projected regret
\[
    R_T
    =
    \sum_{t=1}^{T}
    \left(
        F(\tilde{\theta}_t)-F(\tilde{\theta}^{\star})
    \right)
\]
satisfies
\[
    R_T
    =
    O\left(
        \sqrt{T\beta_T\gamma_T^{(N)}}
    \right),
\]
where $\gamma_T^{(N)}$ is the maximum information gain after $T$ observations under the corresponding heteroscedastic observation uncertainties. Consequently, whenever $\beta_T\gamma_T^{(N)}=o(T)$, the average projected regret $R_T/T$ vanishes.
\end{proposition}
\begin{proof}
The proof follows the standard GP-UCB/LCB regret argument~\citep{srinivas2010gaussian}, applied to the minimization of the projected objective $F$. On the confidence event, for the
selected point $\tilde{\theta}_t$,
\[
    F(\tilde{\theta}_t)
    \leq
    \mu_{t-1}(\tilde{\theta}_t)
    +
    \sqrt{\beta_t}\sigma_{t-1}(\tilde{\theta}_t),
\]
while for the optimal projected point $\tilde{\theta}^{\star}$,
\[
    \mu_{t-1}(\tilde{\theta}^{\star})
    -
    \sqrt{\beta_t}\sigma_{t-1}(\tilde{\theta}^{\star})
    \leq
    F(\tilde{\theta}^{\star}).
\]
Since $\tilde{\theta}_t$ minimizes the lower confidence bound,
\[
    \mu_{t-1}(\tilde{\theta}_t)
    -
    \sqrt{\beta_t}\sigma_{t-1}(\tilde{\theta}_t)
    \leq
    \mu_{t-1}(\tilde{\theta}^{\star})
    -
    \sqrt{\beta_t}\sigma_{t-1}(\tilde{\theta}^{\star}).
\]
Combining the three inequalities gives
\[
    F(\tilde{\theta}_t)-F(\tilde{\theta}^{\star})
    \leq
    2\sqrt{\beta_t}\sigma_{t-1}(\tilde{\theta}_t).
\]
Summing over $t=1,\ldots,T$ and applying the Cauchy--Schwarz inequality yields
\[
\begin{aligned}
    R_T
    &\leq
    2\sum_{t=1}^{T}
    \sqrt{\beta_t}\sigma_{t-1}(\tilde{\theta}_t) \\
    &\leq
    2\sqrt{
        T\beta_T
        \sum_{t=1}^{T}\sigma_{t-1}^2(\tilde{\theta}_t)
    }.
\end{aligned}
\]
The standard information-gain argument for GP bandits bounds the cumulative posterior variance by a constant multiple of $\gamma_T^{(N)}$, yielding
\[
    R_T
    =
    O\left(
        \sqrt{T\beta_T\gamma_T^{(N)}}
    \right).
\]
Dividing by $T$ gives vanishing average projected regret whenever
$\beta_T\gamma_T^{(N)}=o(T)$.
\end{proof}

\emph{Remark. }
Proposition~\ref{prop:idealized_gp_lcb} should be interpreted as an idealized analysis of RuleSmith's continuous-proposal, discrete-evaluation procedure. The optimizer proposes $\tilde{\theta}\in\widetilde{\Theta}$, but the environment evaluates only the projected ruleset $D(\tilde{\theta})$. Consequently, the objective observed by the optimizer is the projected loss
\[
    F(\tilde{\theta}) = \mathcal{L}(D(\tilde{\theta})).
\]
This objective is generally piecewise constant and can be discontinuous at projection-cell boundaries, so the GP calibration assumption is not guaranteed solely by choosing a smooth kernel such as Matérn $5/2$. Rather, it is a standard modeling idealization used to provide theoretical context for GP-based search under noisy self-play evaluations.

The proposition analyzes an LCB rule, whereas the implemented system uses Expected Improvement together with adaptive replication. The result is therefore not a regret guarantee for the implementation. LCB carries the bound because the classical argument~\citep{srinivas2010gaussian} requires an explicit confidence-bound exploration term: the objective is sandwiched by the confidence interval, per-step regret is bounded by the posterior standard deviation at the queried point, and the sum is controlled by the maximum information gain. EI has no comparably standard cumulative-regret bound; its classical guarantee is a convergence rate in the noiseless case~\citep{bull2011convergence}, and results for noisy EI do not cover a discrete projection combined with EI-dependent replication. The two rules nonetheless use the same posterior quantities. Writing $\operatorname{EI}=\sigma\,h(z)$ with $z=(y^*-\mu)/\sigma$ and $h(z)=z\Phi(z)+\phi(z)$, the partial derivatives of EI with respect to the predicted improvement $y^*-\mu$ and $\sigma$ are $\Phi(z)$ and $\phi(z)$, respectively; because both are positive, EI increases in the exact two quantities that LCB trades off, and the rules differ in how they weight them. Lemma~\ref{lemma:concentration_of_loss} justifies the concentration of the self-play estimates, Proposition~\ref{prop:idealized_gp_lcb} shows that GP-based search is well posed for this objective under idealized calibration, and the choice among acquisition rules is settled empirically, including the LCB run in Section~\ref{subsec:statistical_results}.

%%%%%%%%%%%%%%%%%%%%%%%%%%%%%%%%%%%%%%%%%%%%%%%%%%%%%%%%%%%%
\section{Validating the Proxy Player}
\label{sec:supp_proxy_validation}

\subsection{LLM Agents vs.\ Random Agent Baseline}
\label{sec:supp_random_baseline}

To establish a lower bound on proxy-player quality, Table~\ref{tab:supp_vs_random} evaluates InternVL3.5-8B and InternVL3.5-2B against a \textbf{random agent} that selects uniformly from all legal actions. The prompted LLM agents win 100\% of games in every listed configuration, regardless of faction. This establishes effective play relative to random action selection and supports their use as proxy players. As with the other agent comparisons, performance reflects the model together with its prompts and strategy guidance.

\begin{table}[h]
\papertablestyle
\caption{\textbf{LLM agent vs.\ random agent.} LLM agents (InternVL3.5-8B and 2B) are evaluated against a uniformly random legal-action agent using parameters optimized by the respective LLM. The prompted LLM agents win 100\% of games in every listed configuration.}
\label{tab:supp_vs_random}
\vspace{2mm}
\centering
\begin{tabular}{llcc}
\toprule
\textbf{Empire} & \textbf{Nomads} & \textbf{Empire\%} & \textbf{Nomads\%} \\
\midrule
InternVL3.5-8B & Random         & 100 & 0   \\
Random         & InternVL3.5-8B & 0   & 100 \\
InternVL3.5-2B & Random         & 100 & 0   \\
Random         & InternVL3.5-2B & 0   & 100 \\
\bottomrule
\end{tabular}
\end{table}

\subsection{LLM Agents vs.\ Rule-Based Agent}
\label{sec:supp_vs_rule}

To provide a stronger proxy-player baseline than random play, we design a \textbf{rule-based agent} that executes fixed greedy heuristics with random tie-breaking, without language understanding.

\paragraph{Empire rule-based agent.}
Farmers always \texttt{GATHER} if possible; otherwise they \texttt{PASS}. Soldiers \texttt{BATTLE} any adjacent enemy (prioritizing the enemy city), or \texttt{MOVE} toward the nearest enemy by Manhattan distance; if multiple enemies are equidistant, one is chosen uniformly at random. The city alternates each turn between \texttt{PRODUCE\_UNIT} (farmer) and \texttt{PRODUCE\_UNIT} (soldier) whenever resources suffice and an adjacent cell is free; otherwise it uses \texttt{PRODUCE\_RESOURCE}.

\paragraph{Nomads rule-based agent.}
Cavalry \texttt{BATTLE} the enemy city if adjacent; otherwise they \texttt{BATTLE} the nearest adjacent enemy unit; otherwise they \texttt{MOVE} toward the enemy city. The city uses \texttt{PRODUCE\_UNIT} (cavalry) when resources allow; otherwise it uses \texttt{PASS}.

These heuristics capture the core asymmetry of each faction: Empire prioritizes economy and incremental defense, whereas Nomads prioritize fast aggression and city destruction.

Table~\ref{tab:supp_vs_rule} reports win rates when LLM agents face the rule-based agent under two parameter sets: parameters optimized with LLM self-play (Table~\ref{tab:ewin_nwin_matrix}) and parameters optimized with rule-based self-play (Table~\ref{tab:supp_random_params}).

LLM-agent win rates range from 54\% to 78\% in Table~\ref{tab:supp_vs_rule}, below their 100\% rate against random agents. The rule-based agent therefore provides meaningful competition for both model sizes. Under LLM-optimized parameters, 8B agents win 78\% as Nomads and 75\% as Empire. Under rule-based-optimized parameters, these rates are 62\% and 73\%, respectively. The latter configuration has cheaper Empire units (4), higher farmer gathering (2), and lower Nomads kill gain (5), compatible with an economy favoring frequent Empire production. These descriptive comparisons show interactions between rules and player policies; they do not isolate the contribution of individual mechanics. Together, the random and scripted baselines support the use of prompted LLM agents as a more capable evaluation population in these settings.

\begin{table}[h]
\papertablestyle
\caption{\textbf{LLM agent vs.\ rule-based agent.} Win rates when LLM agents (InternVL3.5-8B and 2B) play against the rule-based agent under parameters optimized via LLM self-play (left) and rule-based self-play (right). All results use 100 evaluation games.}
\label{tab:supp_vs_rule}
\vspace{2mm}
\centering
\begin{tabular}{ll cc@{\hspace{10pt}} cc}
\toprule
& & \multicolumn{2}{c}{\textbf{LLM-optimized params}} & \multicolumn{2}{c}{\textbf{Rule-optimized params}} \\
\cmidrule(lr){3-4}\cmidrule(lr){5-6}
\textbf{Empire} & \textbf{Nomads} & \textbf{Empire\%} & \textbf{Nomads\%} & \textbf{Empire\%} & \textbf{Nomads\%} \\
\midrule
InternVL3.5-8B & Rule-based     & 75 & 25 & 73 & 27 \\
Rule-based     & InternVL3.5-8B & 22 & 78 & 38 & 62 \\
InternVL3.5-2B & Rule-based     & 68 & 32 & 65 & 35 \\
Rule-based     & InternVL3.5-2B & 31 & 69 & 46 & 54 \\
\bottomrule
\end{tabular}
\end{table}

\subsection{Comparison with Human-Designed Parameters}
\label{sec:supp_human_vs_opt}

Table~\ref{tab:supp_human_vs_opt} compares the authors' initial configuration (``Initial~$\theta$'') against the configuration discovered by RuleSmith using InternVL3.5-8B. This starting point produces a completely one-sided outcome in its original evaluation (Empire~100\%~$\mid$~Nomads~0\%, loss 1.00). Its low Nomads damage (1) relative to Empire damage (4) and Soldier HP (13) plausibly contributes to this advantage. This interpretation is descriptive; joint parameter changes do not isolate a causal effect for any one coordinate.

RuleSmith's Bayesian optimizer jointly adjusts all parameter categories to recover empirical balance. Nomads' damage triples (1~$\to$~3), their production cost rises (3~$\to$~8), initial resources increase (6~$\to$~10), and Empire Soldier HP falls (13~$\to$~9). These changes are consistent with stronger Nomads combat, costlier cavalry production, and greater initial spending capacity. The optimized combat statistics imply three attacks to eliminate either faction's main combat unit. The empirical balance loss drops from 1.00 to 0.02; these mechanistic interpretations concern the joint configuration rather than isolated parameter effects.

\begin{table}[h]
\papertablestyle
\caption{\textbf{Human-designed vs.\ RuleSmith-optimized parameters.} The authors' initial configuration is heavily skewed toward Empire (balance score 1.00). RuleSmith reduces the balance score to 0.02 through coordinated adjustments across all parameter categories.}
\label{tab:supp_human_vs_opt}
\vspace{2mm}
\centering
\begin{tabular}{lcc@{\hspace{12pt}}lcc}
\toprule
\textbf{Parameter} & \textbf{Initial} & \textbf{Opt.} &
\textbf{Parameter} & \textbf{Initial} & \textbf{Opt.} \\
\midrule
\multicolumn{3}{l}{\emph{Economy}} & \multicolumn{3}{l}{\emph{Production}} \\
Initial resources         & 6            & 10           & Empire unit cost          & 7            & 9            \\
Empire farmer gather      & 2            & 1            & Nomads unit cost          & 3            & 8            \\
Nomads kill resource gain & 4            & 7            &                           &              &              \\
\midrule
\multicolumn{3}{l}{\emph{Combat}} & \multicolumn{3}{l}{\emph{Scoring}} \\
Empire damage             & 4            & 4            & Score per resource        & 0.1          & 0.1          \\
Nomads damage             & 1            & 3            & Score per battle won      & 4            & 3            \\
Empire soldier HP         & 13           & 9            & Score per surviving unit  & 4            & 5            \\
Nomads cavalry HP         & 10           & 12           &                           &              &              \\
\midrule
\multicolumn{3}{l}{\textbf{Empire win rate}: 100\% $\rightarrow$ 51\%} &
\multicolumn{3}{l}{\textbf{Nomads win rate}: 0\% $\rightarrow$ 49\%} \\
\multicolumn{6}{l}{\textbf{Balance score} $\mathcal{L}$: 1.00 $\rightarrow$ 0.02} \\
\bottomrule
\end{tabular}
\end{table}

%%%%%%%%%%%%%%%%%%%%%%%%%%%%%%%%%%%%%%%%%%%%%%%%%%%%%%%%%%%%
\section{Generalization of Balanced Configurations}
\label{sec:supp_generalization}

A central question for RuleSmith is whether balanced game parameters reflect the structural properties of the game itself or are merely artifacts of a specific LLM's strategy distribution. We present five complementary experiments to answer this question.

\subsection{Cross-LLM-Family Transfer}
\label{subsec:supp_cross_llm}

We optimize game parameters using either InternVL3.5-8B or Qwen3.5-9B as the proxy player and then evaluate them with both models. As reported in Table~\ref{tab:cross_llm} (main paper), all four combinations remain near-balanced (balance score $\leq 0.04$), providing evidence of transfer across these LLM families at the tested sample size.

\subsection{Evaluation with Gemini Flash and Pro}
\label{subsec:supp_system12}

Table~\ref{tab:supp_system12} evaluates configurations with \textbf{Gemini~2.5~Flash} and \textbf{Gemini~2.5~Pro}. Parameters optimized with InternVL3.5-8B or Qwen3.5-9B meet the empirical near-balance criterion under both evaluators. This provides evidence of cross-model transfer at the tested sample size; comparing distinct models does not isolate the effect of a reasoning mode.

\begin{table}[h]
\papertablestyle
\caption{\textbf{Evaluation with Gemini Flash and Pro.} Parameters optimized with InternVL3.5-8B or Qwen3.5-9B meet the empirical near-balance criterion when evaluated with Gemini~2.5~Flash or Pro. The comparison varies model identity and does not isolate a reasoning mode.}
\label{tab:supp_system12}
\vspace{2mm}
\centering
\begin{tabular}{llccc}
\toprule
\textbf{Optimization LLM} & \textbf{Evaluation LLM} & \textbf{Empire\%} & \textbf{Nomads\%} & $\mathcal{L}$ \\
\midrule
InternVL3.5-8B & Gemini~2.5~Flash & 51 & 49 & 0.02 \\
InternVL3.5-8B & Gemini~2.5~Pro   & 47 & 53 & 0.06 \\
Qwen3.5-9B     & Gemini~2.5~Flash & 45 & 55 & 0.10 \\
Qwen3.5-9B     & Gemini~2.5~Pro   & 53 & 47 & 0.06 \\
\bottomrule
\end{tabular}
\end{table}

\subsection{Asymmetric Cross-Model Matchups}
\label{subsec:supp_asymmetric}

Table~\ref{tab:supp_asymmetric} reports \emph{asymmetric} matchups using parameters optimized with symmetric InternVL3.5-8B self-play, placing a different LLM on each side. With InternVL3.5-8B playing Empire, Gemini~2.5~Pro as Nomads wins 59\% versus 55\% for Flash; these single evaluations do not establish a statistically reliable ordering. InternVL3.5-8B obtains 41--55\% win rates across the listed matchups, depending on its faction and opponent. These results show that outcomes remain sensitive to the player pairing, so transfer is a property to measure rather than assume.

\begin{table}[h]
\papertablestyle
\caption{\textbf{Asymmetric cross-model matchups.} Parameters optimized with symmetric InternVL3.5-8B self-play are evaluated with different models on the two sides. Faction outcomes vary with the pairing, while both factions retain wins in the tested matchups.}
\label{tab:supp_asymmetric}
\vspace{2mm}
\centering
\begin{tabular}{llccc}
\toprule
\textbf{Empire} & \textbf{Nomads} & \textbf{Empire\%} & \textbf{Nomads\%} & $\mathcal{L}$ \\
\midrule
InternVL3.5-8B  & Gemini~2.5~Flash & 45          & 55          & 0.10 \\
InternVL3.5-8B  & Gemini~2.5~Pro   & 41          & 59          & 0.18 \\
Gemini~2.5~Flash & InternVL3.5-8B  & 45          & 55          & 0.10 \\
Gemini~2.5~Pro  & InternVL3.5-8B   & 49 & 51 & 0.02 \\
\bottomrule
\end{tabular}
\end{table}

\subsection{Human User Study}
\label{sec:supp_user_study}

To evaluate whether the game parameters balanced via LLM self-play also produce fair outcomes for human players, we conducted a small-scale user study using the parameters optimized by RuleSmith with InternVL3.5-8B.

\paragraph{Setup.}
Four of the authors (graduate students with prior experience in turn-based strategy games) took part. We arranged all $\binom{4}{2} = 6$ unique pairwise matchups; each pair played with both faction assignments (Empire vs.\ Nomads, then Nomads vs.\ Empire), yielding 12 game sessions in total. Each session consisted of 5 games, for a total of \textbf{60 games}. Participants received the same natural-language rulebook provided to the LLM agents, with no additional strategic guidance. We recorded only aggregate win/loss outcomes and did not collect personal or sensitive data.

\paragraph{Results.}
Table~\ref{tab:supp_user_study} summarizes the outcomes. Across 60 games, Empire won 32 times and Nomads won 28 times (53.3\% vs.\ 46.7\%), with no draws. Both individual win rates (53.3\% and 46.7\%) fall within the $50\%\pm5\%$ near-balanced range used as our criterion throughout this work.

\begin{table}[h]
\papertablestyle
\caption{\textbf{Human user study results.} 4 participants, 6 pairwise matchups, 2 faction assignments each, 5 games per session, 60 games total. Empire win rate 53.3\% and Nomads win rate 46.7\% both fall within the near-balanced range ($50\%\pm5\%$).}
\label{tab:supp_user_study}
\vspace{2mm}
\centering
\begin{tabular}{lccc}
\toprule
 & \textbf{Empire} & \textbf{Nomads} & \textbf{Total} \\
\midrule
\textbf{Wins}     & 32      & 28     & 60  \\
\textbf{Win Rate} & 53.3\%  & 46.7\% & N/A \\
\bottomrule
\end{tabular}
\end{table}

These results provide preliminary evidence of transfer to these human participants; the 60 games do not establish equivalence within five percentage points or population-wide human balance. We acknowledge that this study is small in scale and that a larger, pre-registered human study would provide stronger evidence; we leave this to future work.

\input{human_agent_study.tex}

%%%%%%%%%%%%%%%%%%%%%%%%%%%%%%%%%%%%%%%%%%%%%%%%%%%%%%%%%%%%
\section{Scalability and Extended Configurations}
\label{sec:supp_extended}

\subsection{Robustness to Map Size and Turn Limit}
\label{subsec:supp_map_size}

Table~\ref{tab:extended} in the main text evaluates RuleSmith across four map sizes with corresponding turn limits (16 turns for 5$\times$5 and 7$\times$7; 32 turns for 9$\times$9 and 11$\times$11). Near-balanced outcomes hold in all settings, providing initial evidence across these game scales.

\subsection{Three-Faction Balancing}
\label{subsec:supp_three_faction}

To evaluate scalability beyond two factions, we introduce a third faction, \emph{Theocracy}, to CivMini. Theocracy produces \texttt{Priest} units that can \texttt{MOVE} and \texttt{CONVERT} adjacent enemy Farmers, Soldiers, or Cavalry over $1$--$5$ game turns (optimizable); Priests sacrifice themselves upon conversion. Four new parameters are added to the search space: Priest production cost, Priest HP, Theocracy city attack damage, and conversion time. The balance loss is extended to the three-faction objective $\mathcal{L}_3(\theta) = |w_E - \nicefrac{1}{3}| + |w_N - \nicefrac{1}{3}| + |w_T - \nicefrac{1}{3}|$, targeting equal win rates among all three factions.

Table~\ref{tab:extended} shows that RuleSmith successfully balances all three factions simultaneously under both InternVL3.5-8B and InternVL3.5-2B, reaching losses of 0.033 and 0.013, respectively, based on the reported faction win rates.

\subsection{Grid Size as a Rule-Level Parameter}
\label{subsec:supp_gridsize}

Beyond numerical tuning, we test whether RuleSmith handles rule-level structural changes by adding the map grid size (integer range $[5, 9]$) as an optimizable parameter alongside the standard 12 numeric parameters. Table~\ref{tab:extended} shows that RuleSmith converges to near-balanced outcomes under both model sizes, demonstrating that the framework naturally extends to structural game decisions, not just fine-grained numerical adjustments.

\subsection{Rule Parameters: LLM Agents, Random Agents, and Rule-Based Agents}
\label{sec:supp_random_params}

Table~\ref{tab:supp_random_params} compares parameters optimized by BO under LLM, random, and rule-based self-play. All configurations are subsequently evaluated with InternVL3.5-8B; mixed LLM/rule-based evaluations are reported separately in Table~\ref{tab:supp_vs_rule}. The distinct configurations illustrate how the choice of proxy population affects the discovered ruleset.

\textbf{LLM-optimized parameters} include initial resources of 10, a kill resource gain of 7, and Nomads HP of 12, compatible with sustained combat and unit replacement. \textbf{Random-agent parameters} use initial resources of 4 and a lower battle-scoring weight, placing relatively more scoring weight on resource inventories. \textbf{Rule-based-agent parameters} feature cheaper Empire units (4 versus 9) and greater Nomads damage (4 versus 3). The latter reduces Nomads' TTK against Empire Soldiers from three attacks to two. Cheaper production and faster combat may help the scripted heuristics remain competitive. These numerical contrasts provide interpretable descriptions of the joint solutions; isolating individual causal effects would require separate parameter ablations.

\begin{table}[h]
\papertablestyle
\caption{\textbf{Optimized parameters: LLM agents, random agents, and rule-based agents.} The three proxy player types yield substantially different balanced configurations, reflecting their distinct decision dynamics.}
\label{tab:supp_random_params}
\vspace{2mm}
\centering
\begin{tabular}{lccc}
\toprule
\textbf{Parameter} & \textbf{LLM Agents} & \textbf{Random Agents} & \textbf{Rule-Based Agents} \\
\midrule
\multicolumn{4}{l}{\emph{Economy}} \\
Initial resources            & 10  & 4   & 6   \\
Empire farmer gather         & 1   & 1   & 2   \\
Nomads kill resource gain    & 7   & 5   & 5   \\
\midrule
\multicolumn{4}{l}{\emph{Combat}} \\
Empire damage                & 4   & 2   & 3   \\
Nomads damage                & 3   & 5   & 4   \\
Empire soldier HP            & 9   & 13  & 8   \\
Nomads cavalry HP            & 12  & 6   & 9   \\
\midrule
\multicolumn{4}{l}{\emph{Production}} \\
Empire unit cost             & 9   & 8   & 4   \\
Nomads unit cost             & 8   & 4   & 5   \\
\midrule
\multicolumn{4}{l}{\emph{Scoring}} \\
Score per resource           & 0.1 & 0.3 & 0.2 \\
Score per battle won         & 3   & 1   & 2   \\
Score per surviving unit     & 5   & 4   & 3   \\
\bottomrule
\end{tabular}
\end{table}

%%%%%%%%%%%%%%%%%%%%%%%%%%%%%%%%%%%%%%%%%%%%%%%%%%%%%%%%%%%%
\section{Supplementary Analysis}
\label{sec:supp_analysis}

\subsection{Optimized Parameters Comparison}
\label{sec:supp_optimized_parameters}

Table~\ref{tab:supp_optimized_parameters} summarizes the parameters that yield near-even empirical win rates under the different LLM pairings. The optimized configurations vary with agent capacity and asymmetry.

For symmetric-capacity pairings (\textit{E}$_{2\mathrm{B}}$ vs \textit{N}$_{2\mathrm{B}}$ and \textit{E}$_{8\mathrm{B}}$ vs \textit{N}$_{8\mathrm{B}}$), the implied time-to-kill (TTK) is three attacks in both directions. These similar attack counts are compatible with competitive combat, while faction mobility and economic roles remain asymmetric. TTK describes the main combat units and does not by itself measure strategic diversity.

For \textit{E}$_{8\mathrm{B}}$ vs \textit{N}$_{2\mathrm{B}}$, Nomads require four attacks to eliminate an Empire Soldier, whereas Empire requires eight to eliminate Cavalry. This favors the lower-capacity Nomads side in direct combat. Empire units also cost more than Nomads units (6 versus 4), and battle and unit scoring weights are lower than in the symmetric 8B configuration. Together, these edits are consistent with compensating for the capacity difference.

For \textit{E}$_{2\mathrm{B}}$ vs \textit{N}$_{8\mathrm{B}}$, TTK still slightly favors Nomads (five versus six attacks), but initial resources fall from 10 in the symmetric 8B setting to 2. This constrains early production for both factions. Higher farmer gathering (2 versus 1) and cheaper Empire units (4 versus 5 for Nomads) may help the lower-capacity Empire remain competitive. These interpretations concern the combined configuration, whose observed outcomes are near-balanced.

These comparisons illustrate coordinated edits across economy, combat, production, and scoring. RuleSmith discovers distinct near-balanced configurations for different player pairings, retaining faction asymmetry while adjusting numerical strength. This supports flexibility across the evaluated populations without assuming a single canonical balanced ruleset.

\begin{table}[h]
\papertablestyle
\caption{\textbf{Optimized game parameters under different LLM agent configurations.} We report optimized parameters that achieve approximately 50/50 win rates across different LLM agent configurations. We also include the implied time-to-kill (TTK; number of attacks required to eliminate the opponent's main combat unit) derived from damage and HP.}
\label{tab:supp_optimized_parameters}
\vspace{2mm}
\centering
\begin{tabular}{lcccc}
\toprule
\textbf{Parameter}
& \textbf{\textit{E}$_{2\mathrm{B}}$ vs \textit{N}$_{2\mathrm{B}}$}
& \textbf{\textit{E}$_{2\mathrm{B}}$ vs \textit{N}$_{8\mathrm{B}}$}
& \textbf{\textit{E}$_{8\mathrm{B}}$ vs \textit{N}$_{2\mathrm{B}}$}
& \textbf{\textit{E}$_{8\mathrm{B}}$ vs \textit{N}$_{8\mathrm{B}}$} \\
\midrule
Initial resources            & 10 & 2  & 10 & 10 \\
Empire farmer gather         & 1  & 2  & 4  & 1  \\
Nomads kill resource gain    & 6  & 9  & 1  & 7  \\
Empire damage                & 4  & 2  & 2  & 4  \\
Nomads damage                & 3  & 3  & 5  & 3  \\
Empire soldier HP            & 9  & 14 & 16 & 9  \\
Nomads cavalry HP            & 10 & 11 & 16 & 12 \\
\midrule
\textbf{TTK$_{\text{N}\rightarrow\text{E}}$ ($\lceil \text{HP}_{E}/\text{DMG}_{N}\rceil$)}
& \textbf{3} & \textbf{5} & \textbf{4} & \textbf{3} \\
\textbf{TTK$_{\text{E}\rightarrow\text{N}}$ ($\lceil \text{HP}_{N}/\text{DMG}_{E}\rceil$)}
& \textbf{3} & \textbf{6} & \textbf{8} & \textbf{3} \\
\midrule
Empire unit cost             & 4  & 4  & 6  & 9  \\
Nomads unit cost             & 4  & 5  & 4  & 8  \\
Score per resource           & 0.4& 0.2& 0.3& 0.1\\
Score per battle won         & 3  & 4  & 1  & 3  \\
Score per surviving unit     & 3  & 4  & 1  & 5  \\
\bottomrule
\end{tabular}
\vspace{-3mm}
\end{table}

\subsection{Representative Rollout}
\label{subsec:representative_rollout}

\begin{figure}[!tb]
    \centering
    \includegraphics[width=\linewidth]{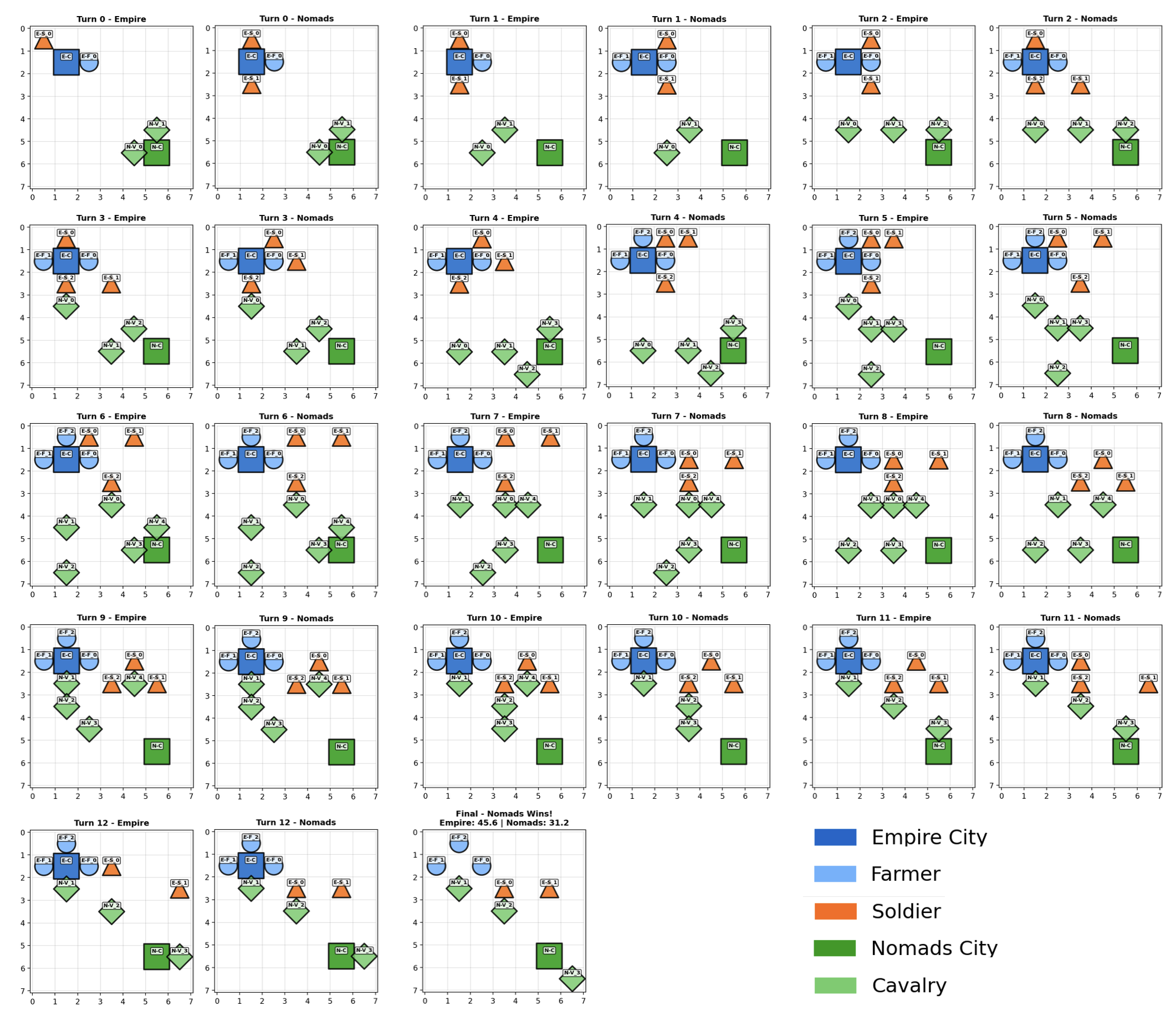}
    \vspace{-3mm}
    \caption{
        \textbf{Visualization of a CivMini Game.} Nomads conquer the Empire's city on turn 12 (13 rounds, counting from 0). Cavalry bypass the Empire's defense and sustain city pressure. Both agents use InternVL3.5-8B. This example is R94 game 13 (Table~\ref{tab:rollout_configurations}); city destruction determines victory irrespective of the displayed scores.}
    \label{fig:game_vis}
    \vspace{-3mm}
\end{figure}

\paragraph{Rollout configurations.}
The qualitative examples use two archived 8B/8B optimization checkpoints: R94 (iteration 94) and R98 (iteration 98). Figure~\ref{fig:game_vis} shows R94 game 13, while Figures~\ref{fig:game_rollout_1}--\ref{fig:game_rollout_4} show R98 games 4, 1, 5, and 13, respectively. Table~\ref{tab:rollout_configurations} lists their parameters alongside the reported 8B/8B configuration. These examples illustrate tactics at selected checkpoints; population-level balance is assessed by the separate held-out evaluations. Turn labels in these figures start at 0, so a game ending on turn 12 spans 13 rounds and one ending on turn 6 spans 7.

\begin{table}[htbp]
\papertablestyle
\centering
\caption{\textbf{Configurations used in the qualitative rollouts.} R94 and R98 are archived optimization checkpoints, distinct from the reported 8B/8B configuration in Table~\ref{tab:supp_optimized_parameters}.}
\label{tab:rollout_configurations}
\begin{tabular}{lrrr}
\toprule
Parameter & R94 & R98 & Reported 8B/8B \\
\midrule
Initial resources & 8 & 4 & 10 \\
Empire farmer gather & 1 & 1 & 1 \\
Nomads kill resource gain & 8 & 7 & 7 \\
Empire damage & 2 & 2 & 4 \\
Nomads damage & 4 & 4 & 3 \\
Empire soldier HP & 14 & 14 & 9 \\
Nomads cavalry HP & 8 & 8 & 12 \\
Empire unit cost & 3 & 6 & 9 \\
Nomads unit cost & 5 & 4 & 8 \\
Score per resource & 0.4 & 0.2 & 0.1 \\
Score per battle won & 3 & 4 & 3 \\
Score per surviving unit & 4 & 4 & 5 \\
\bottomrule
\end{tabular}
\end{table}

For R94 and R98, the battle multiplier is 1 and each city produces 3 resources per \texttt{PRODUCE\_RESOURCE} action. In addition to their cities, Empire starts with one Farmer and one Soldier, and Nomads start with two Cavalry.

Figure~\ref{fig:game_vis} shows a Nomads conquest on turn 12. Cavalry mobility allows a unit to bypass the Empire's central formation and reach its city. The Empire redeploys Soldiers to intercept, but sustained city pressure ends the game. This illustrates how a narrow opening can become a decisive conquest even when the Empire retains units elsewhere on the board.

\subsection{Extra Visualization for Game Rollouts}
\label{sec:supp_vis}

\begin{figure}[h!]
    \centering
    \includegraphics[width=0.8\linewidth, keepaspectratio]{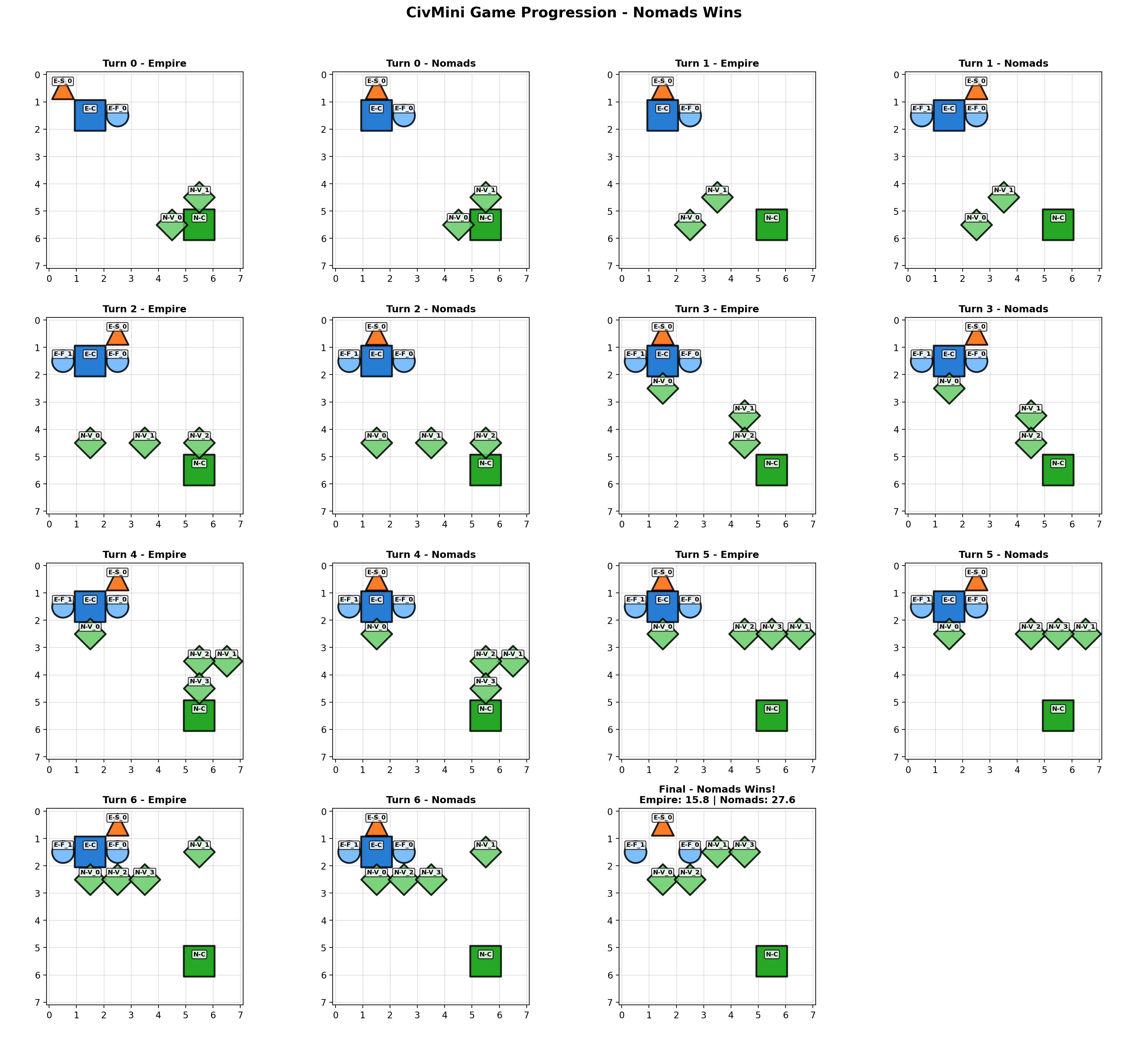}
    \vspace{-3mm}
    \caption{A rapid Nomads conquest on turn 6 (7 rounds, counting from 0). Cavalry Unit 0 strikes the Empire's city on turns 3--6, dealing 4 HP per attack and destroying it. Checkpoint R98, game 4. (Final score: Empire 15.8 | Nomads 27.6)}
    \label{fig:game_rollout_1}
    \vspace{-5mm}
\end{figure}

\begin{figure}[h!]
    \centering
    \includegraphics[height=0.86\textheight, width=\linewidth, keepaspectratio]{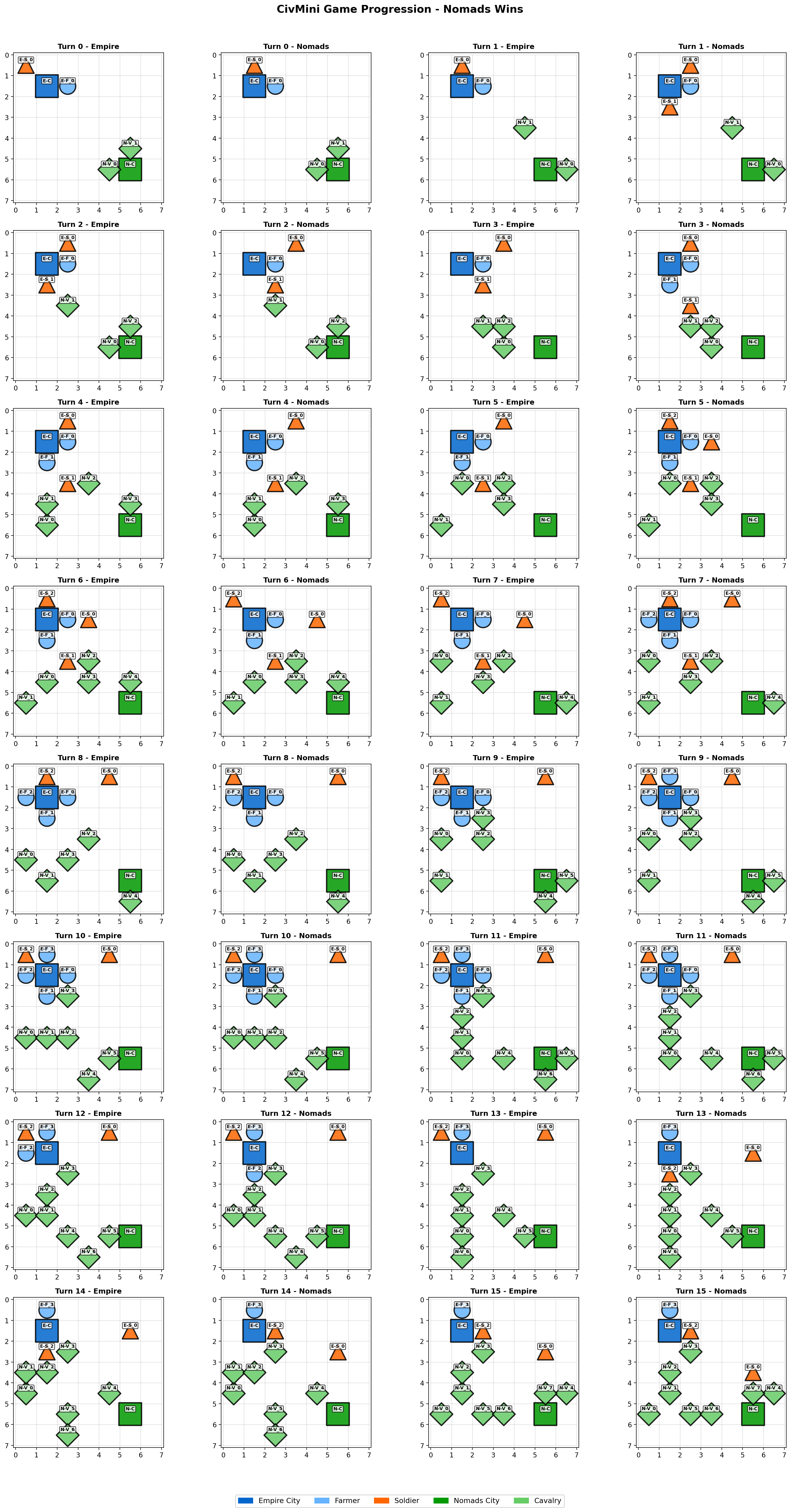}
    \caption{A Nomads score victory after 16 rounds (turns 0--15). Continued cavalry production builds a numerical advantage, while the Empire deploys Farmers and Soldiers and loses units during the fighting. Checkpoint R98, game 1. (Final score: Empire 14.2 | Nomads 67.6)}
    \label{fig:game_rollout_2}
\end{figure}

\begin{figure}[h!]
    \centering
    \includegraphics[height=0.86\textheight, width=\linewidth, keepaspectratio]{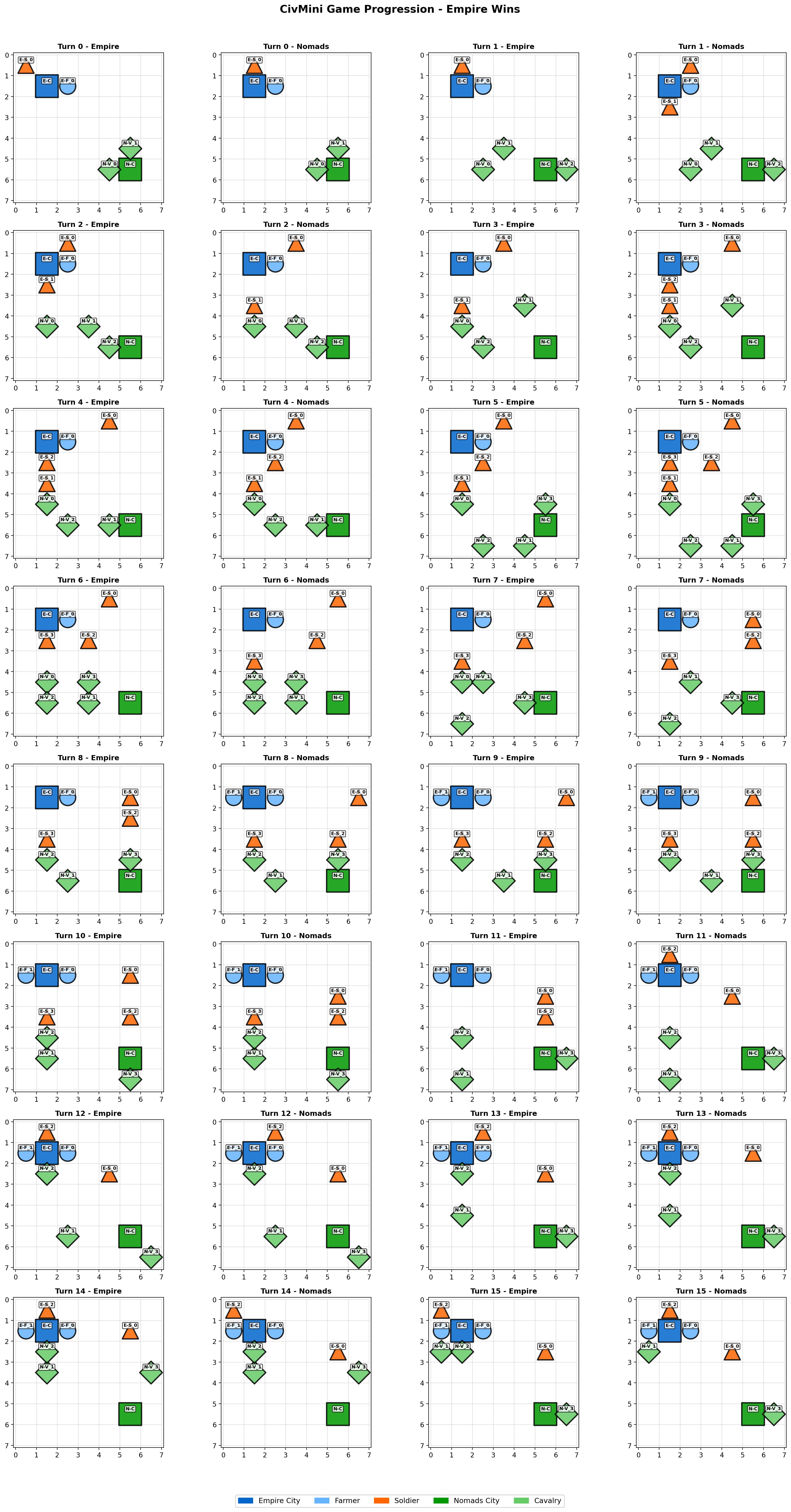}
    \caption{The Empire withstands early Nomads aggression and wins at the 16-round turn cap. Soldiers defend through fighting on turns 3--15, while Farmers supply resources. Two Cavalry remain at the end; the Empire wins through the combined final scoring terms. Checkpoint R98, game 5. (Final score: Empire 31.6 | Nomads 23.4)}
    \label{fig:game_rollout_3}
\end{figure}

\begin{figure}[h!]
    \centering
    \includegraphics[height=0.86\textheight, width=\linewidth, keepaspectratio]{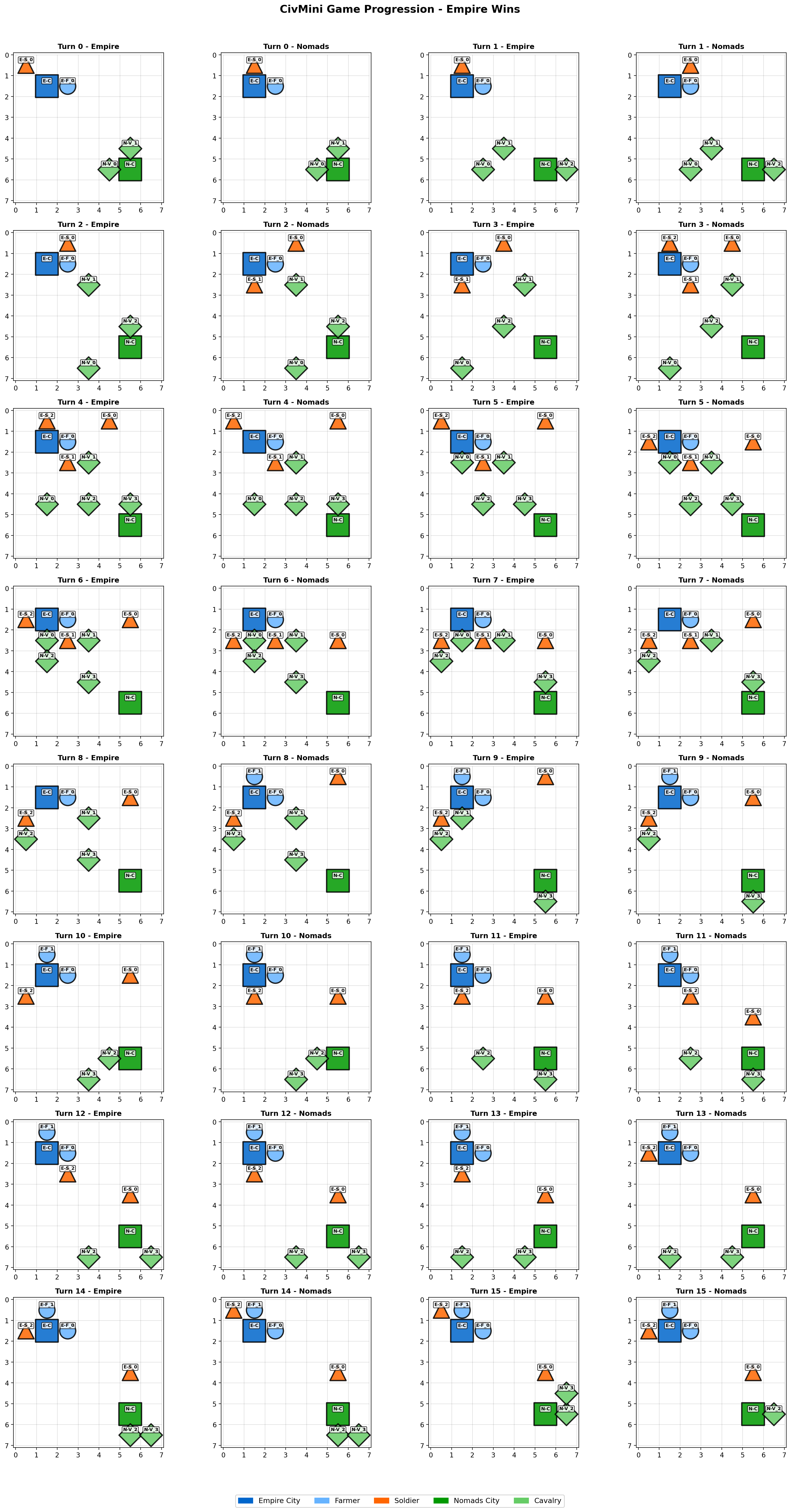}
    \caption{The Empire defends against a Nomads city rush beginning on turn 5. By turn 15, one Cavalry remains, while the Empire retains two Soldiers and two Farmers. Unit survival and resources contribute to its score victory. Checkpoint R98, game 13. (Final score: Empire 34.4 | Nomads 12.6)}
    \label{fig:game_rollout_4}
\end{figure}

We provide additional qualitative visualizations with InternVL3.5-8B agents on both sides under checkpoint R98 (Table~\ref{tab:rollout_configurations}). Figures~\ref{fig:game_rollout_1}--\ref{fig:game_rollout_4} show both factions winning through distinct tactical paths, illustrating early aggression, unit composition, and economic decisions under identical rules.

Figure~\ref{fig:game_rollout_1} shows a fast Nomads victory. Cavalry Unit 0 repeatedly attacks the Empire's city from turn 3 through turn 6, dealing 4 HP per strike. Four attacks destroy the 14-HP city, ending the game in its seventh round. This example illustrates a concentrated city attack succeeding before local defenses can stop it.

In contrast, Figure~\ref{fig:game_rollout_2} shows a longer game in which the Nomads secure a high-score victory through sustained expansion and force dominance. By continuously producing cavalry units, the Nomads overwhelm the Empire's defensive formation. Although the Empire attempts to stabilize the board with farmers and soldiers, its inability to eliminate cavalry units leads to gradual territorial collapse and a board state fully dominated by Nomad forces.

Figures~\ref{fig:game_rollout_3} and~\ref{fig:game_rollout_4} show the Empire countering early Nomads aggression. In Figure~\ref{fig:game_rollout_3}, Soldiers hold a defensive line through a prolonged engagement, while Farmers generate resources. The Empire wins at the turn cap through its final resources, battle outcomes, and surviving units.

Similarly, Figure~\ref{fig:game_rollout_4} shows a failed Nomad city rush. Although the Nomads initiate a direct attack on the Empire's city as early as turn 5, the Empire's soldiers systematically eliminate the attacking cavalry units. As the Nomad force diminishes, the Empire maintains both a military presence and economic growth, ultimately winning by preserving units and accumulating resources.

These selected rollouts illustrate different tactical paths under the optimized rules: the Nomads win through city pressure or numerical advantage, while Empire victories combine defense, resource production, and unit survival. They complement the larger trajectory diagnostics in Appendix~\ref{sec:supp_agent_diagnostics}. These examples demonstrate observed behavioral variety; establishing a causal effect of optimization on strategic diversity or player enjoyment would require a controlled comparison.

\clearpage
\input{revision_appendix.tex}

\ifarxiv\else
\input{checklist.tex}
\fi

\end{document}

%% file: tables/statistical_evaluation.tex
\begin{table}[!htbp]
\papertablestyle
\centering\caption{\textbf{Additional held-out evaluations.} Empire win rates, Wilson 95\% intervals, and two-sided exact binomial tests against 0.5. Counts are drawn from the evaluation logs; the test compares Empire and Nomads wins among decisive games (Appendix~\ref{sec:supp_statistics}). The default batch is separate from the authors' initial configuration. NTBEA/TPE replace the search component; GP-LCB replaces the acquisition rule.}
\label{tab:statistical_evaluation}
\begin{tabular}{lrrrr}
\toprule
Configuration & Games & Empire (\%) & 95\% CI (\%) & $p$ \\
\midrule
Default (additional evaluation) & 300 & 11.3 & [8.2, 15.4] & $2.9\times10^{-45}$ \\
Optimized configuration (Table~\ref{tab:supp_human_vs_opt}) & 500 & 47.6 & [43.3, 52.0] & 0.30 \\
GP run 1 & 300 & 51.7 & [46.0, 57.3] & 0.60 \\
GP run 2 & 300 & 49.3 & [43.7, 55.0] & 0.86 \\
GP run 3 & 300 & 52.7 & [47.0, 58.2] & 0.39 \\
NTBEA & 300 & 48.0 & [42.4, 53.6] & 0.53 \\
TPE & 300 & 54.3 & [48.7, 59.9] & 0.15 \\
GP-LCB (single run) & 300 & 57.7 & [52.0, 63.1] & 0.009 \\
\bottomrule
\end{tabular}
\end{table}

%% file: human_agent_study.tex
\subsection{Human versus LLM Agents}
\label{subsec:human_vs_agent}
Three of the authors each played 40 games under the configuration optimized with InternVL3.5-8B, giving 120 games in total. Each participant played ten games in each of the four combinations of opponent scale (2B/8B) and human faction (Empire/Nomads) in a randomized order, without knowing which opponent was faced. Randomizing the order spreads participant learning evenly across the conditions, and blinding removes expectation effects from knowing the opponent's size. Per-participant outcomes are reported in Table~\ref{tab:human_agent}.

\begin{table}[htbp]
\papertablestyle
\centering\caption{Human wins by participant and faction against InternVL3.5 agents. Each participant/faction cell has 10 games; totals pool the three participants.}
\label{tab:human_agent}
\begin{tabular}{llrr}
\toprule
Participant & Human faction & vs. 2B & vs. 8B \\
\midrule
P1 & Empire & 9/10 & 8/10 \\
P1 & Nomads & 10/10 & 8/10 \\
P2 & Empire & 8/10 & 7/10 \\
P2 & Nomads & 9/10 & 9/10 \\
P3 & Empire & 9/10 & 7/10 \\
P3 & Nomads & 10/10 & 9/10 \\
\midrule
Total human wins & & 55/60 & 48/60 \\
Human wins as Empire & & 26/30 & 22/30 \\
Human wins as Nomads & & 29/30 & 26/30 \\
\bottomrule
\end{tabular}
\end{table}

Humans win 91.7\% against 2B agents and 80.0\% against 8B, with nominal Wilson 95\% intervals of [81.9, 96.4]\% and [68.2, 88.2]\%. As Empire, humans win 26/30 and 22/30; as Nomads, they win 29/30 and 26/30, respectively. Agent wins increase from 8.3\% to 20.0\% with model scale, a descriptive ordering consistent with the lower execution-conflict rate of the larger model. These outcomes reflect a substantial human--agent skill gap in both faction assignments.

\begin{table}[htbp]
\papertablestyle
\centering\caption{\textbf{Aggregate faction outcomes across small human-player studies.} Wilson intervals and binomial $p$ values summarize game-level totals under a working independence assumption. Repeated participants and learning limit population-level interpretation.}
\label{tab:human_faction}
\begin{tabular}{lrrrr}
\toprule
Population & Empire wins & E (\%) & E 95\% CI (\%) & $p$ \\
\midrule
Human vs. human & 32/60 & 53.3 & [40.9, 65.4] & 0.70 \\
Human vs. 2B & 27/60 & 45.0 & [33.1, 57.5] & 0.52 \\
Human vs. 8B & 26/60 & 43.3 & [31.6, 55.9] & 0.37 \\
\bottomrule
\end{tabular}
\end{table}

Each participant wins 17--19/20 against 2B and 16/20 against 8B. Most agent wins came in a participant's first games, before the participant attended to the scoring rules, and were decided on points at final scoring. Faction outcomes stay close to even in every population across a wide range of playing strength (Table~\ref{tab:human_faction}): Empire wins 45.0\% against the 2B agent ($p=0.52$), 43.3\% against the 8B agent ($p=0.37$), and 53.3\% between humans ($p=0.70$), with none deviating detectably from an even split. Larger studies with more participants would sharpen these estimates.

\paragraph{Participants, instructions, and consent.}
All participants in both human studies were authors of this paper. They took part voluntarily and without compensation, played through a text interface showing the game state, and received the natural-language rulebook (Appendix~\ref{sec:supp_reproducibility}) as their only instructions; no strategic guidance was provided. Only game outcomes were recorded, and no personal or identifying data were collected. Because the participants were the investigators themselves and the sessions posed no risk beyond ordinary gameplay, the study was not submitted for institutional review.

%% file: revision_appendix.tex
\section{Additional Statistical Evaluations}
\label{sec:supp_statistics}
The additional evaluations include three independent GP optimization runs, each evaluated on separate 200- and 300-game batches, and one selected configuration each from NTBEA, TPE, and a GP-LCB run, evaluated at the same two sample sizes. The originally reported optimized configuration is also evaluated on 500 games. These are final-configuration evaluations, not per-iteration confidence intervals or comparisons of convergence speed. Table~\ref{tab:all_reevaluations} reports every additional 200-/300-game batch.

\input{tables/all_reevaluations.tex}

\paragraph{Interval and test definitions.}
For $k$ Empire wins in $n$ games, write $\hat p=k/n$ and $z=1.9599639845$. The Wilson interval is
\[
\frac{\hat p+z^2/(2n)\;\pm\;z\sqrt{\hat p(1-\hat p)/n+z^2/(4n^2)}}{1+z^2/n}.
\]
For no-draw games, a two-sided exact binomial test of $p_E=0.5$ uses
\[
 p_{\mathrm{test}}=\min\!\left\{1,\;2\sum_{j=0}^{\min(k,n-k)} \binom{n}{j}2^{-n}\right\}.
\]
Counts are taken from the evaluation logs. When a batch contains draws, the test compares Empire and Nomads wins among decisive games, and draws are reported separately; the marginal Empire interval uses all games.

The sample SD of the loss across the three GP runs is computed with denominator $3-1$, separately for each evaluation size. It includes optimization-run and evaluation variability; it is not an error bar on every iteration of the optimization trajectory. No multiple-comparison-adjusted claim of superiority is made. Failure to reject an even split does not establish equivalence within $[0.45,0.55]$. Bayesian analyses calibrated to frequentist operating characteristics~\citep{liu2025calibrated,liu2026stepped} are an alternative for small evaluation batches; we report frequentist intervals and exact tests for transparency.

\paragraph{Finite-sample loss.}
Because the loss contains absolute values, its empirical estimate has finite-sample bias as well as sampling variance, both of which depend on the game count. Selected configurations are therefore assessed on independent held-out batches at fixed reported sample sizes, separately from the losses observed during adaptive-budget optimization. Choosing how many games to spend on an evaluation is a sample-size design question, as in clinical trial design~\citep{owen2025comparison}, and allocating games as evidence accumulates during optimization parallels group sequential designs~\citep{liu2023group}; the held-out evaluations use sample sizes fixed in advance.

\input{tables/original_intervals.tex}

\paragraph{Starting configurations.}
The default configuration in the released code sets initial resources to 5, Farmer gather to 2, Nomads kill gain to 3, damage to 1 (Empire) and 2 (Nomads), HP to 8 (Soldier) and 6 (Cavalry), unit cost to 4 for both factions, and score weights to 0.1, 1, and 1. Its 300-game evaluation with 8B agents gives 34 Empire wins, 264 Nomads wins, and 2 draws (Empire 11.3\%, Wilson interval [8.2, 15.4]\%; exact binomial test of Empire against Nomads wins among the 298 decisive games $p=2.9\times10^{-45}$). The qualitative rollout checkpoints R94/R98 are identified separately in Table~\ref{tab:rollout_configurations}.

\section{Agent Validity and Trajectory Diversity}
\label{sec:supp_agent_diagnostics}

\subsection{Execution-Level Conflicts}
All accepted state transitions obey the engine's rules. A joint action proposal can still fail at execution when an earlier action changes its destination occupancy or target. The statistics recorded below count attempted unit actions converted to \texttt{PASS} at execution; they are not a comprehensive rate of unparseable model outputs or hallucinated rule statements.

\begin{table}[htbp]
\papertablestyle
\centering\caption{Execution-level rejection statistics from additional evaluations.}
\label{tab:execution_conflicts}
\begin{tabular}{lrrr}
\toprule
Setting & Attempted actions & Converted to \texttt{PASS} & Rate \\
\midrule
8B agents, 500 games & 33,834 & 1,126 & 3.33\% \\
2B agents, 200 games & 21,871 & 1,012 & 4.63\% \\
\bottomrule
\end{tabular}
\end{table}

In the 8B evaluation, 963 rejections involve movement and 163 involve battle; most result from within-turn resolution conflicts. The 2B rejections are mostly movement conflicts. These rates show a capability-dependent burden from joint action planning, but do not isolate model scale from every other source of variation in these two evaluation batches. Legality enforcement also cannot detect a legal but strategically poor action.

\subsection{Trajectory Diversity}
We analyze the 300-game evaluation of the NTBEA configuration with 8B agents (Empire 48.0\%) and its first 100 games. The analysis uses the executed per-unit actions. Openings are the sequence of all actions by both factions in turns 0--2; entropy is $-\sum_a p(a)\log_2 p(a)$ over action types per faction; board divergence is the mean pairwise Jaccard distance $1-|A\cap B|/|A\cup B|$ between the sets of (cell, unit type) pairs after the Nomads' turn~5 (or the last recorded state for shorter games). The analysis script is included in the release, and the results are in Table~\ref{tab:trajectory_diversity} of the main text.

The first turn offers few legal action combinations, but trajectories diverge over subsequent decisions; the most frequent three-turn opening accounts for 15 of the 300 games. Games end both by conquest and at the turn cap. The diagnostics show that evaluation games do not replay a single line; they do not establish strategic optimality or coverage of the strategy space.

\section{Implementation, Prompts, and Settings}
\label{sec:supp_reproducibility}
The MIT-licensed repository \url{https://github.com/Adonis-galaxy/RuleSmith} provides the game engine, prompts, self-play, the GP, ES, random-search, NTBEA, and TPE optimizers, evaluation, and visualization.

\subsection{Settings}
\begin{table}[htbp]
\papertablestyle
\centering
\caption{Settings for the additional runs (Sections~\ref{subsec:statistical_results} and~\ref{sec:ablation}). The main runs used the same game and agent settings.}
\label{tab:settings}
\begin{tabular}{p{0.28\linewidth}p{0.64\linewidth}}
\toprule
Component & Setting \\
\midrule
Agents & InternVL3.5-2B/8B, FP16, one call per player turn, at most 300 new tokens, temperature 0.7, top-$p$ 0.9, no regeneration; rule checker disabled \\
Retrieval & TF-IDF with cosine similarity over the 10 rulebook entries, query ``multi-unit actions turn structure'', top 3 \\
Prompt & Up to 5 own units listed per decision \\
Game & $7\times7$ map, 16-turn cap, Farmer HP 5, city HP equal to the faction's combat-unit HP, battle multiplier 1, city production 3 per action, movement 1 (Empire) and 2 (Cavalry), Empire starts with one Farmer and one Soldier and Nomads with two Cavalry \\
GP optimizer & scikit-optimize 0.10.2: Gaussian process with constant amplitude times ARD Mat\'ern-5/2 kernel, Gaussian noise, normalized targets; 10 random initial points; acquisition optimized with L-BFGS; EI ($\xi=0.01$); LCB runs use $\kappa=1.96$ \\
Adaptive budget & Eq.~\eqref{eq:adaptive-games}, $N_{\min}=16$, $N_{\max}=64$, $\epsilon=10^{-3}$, integer truncation \\
Final selection & Configuration with the lowest observed in-loop loss (ties: earliest) \\
GP runs & Seeds 1, 2, 3 for both the BO random state and the per-game seed base (game $i$ of iteration $t$ uses seed $10^5 s+100t+i$) \\
LCB run & Seed 1, otherwise identical to GP run~1, so the two share their initial design; 100 iterations (2,899 games) \\
NTBEA & Seed 11; 1-tuples, 2-tuples, and the full 12-tuple over the discretized grid; UCB constant $C=1$; 50 mutated neighbors per step; 68 evaluations of 16 games (1,088 games); uniformly random start \\
TPE & Optuna 4.9.0~\citep{akiba2019optuna} \texttt{TPESampler} defaults, seed 12; 68 trials of 16 games (1,088 games) \\
Held-out evaluation & Per-game seed bases disjoint from in-loop seeds: NTBEA 311 and 611, TPE 312 and 612, GP-LCB 901 and 1001 (200 and 300 games), default configuration 1100; the batches behind Table~\ref{tab:execution_conflicts} use 1000 (8B) and 1200 (2B) \\
Software & PyTorch 2.6.0, Transformers 4.57.6, scikit-learn 1.8.0 \\
\bottomrule
\end{tabular}
\end{table}

\input{appendix_prompts.tex}

\subsection{Models and Figure Tools}
InternVL3.5~\citep{internvl3.5} (2B/8B) and Qwen3.5-9B are released under Apache-2.0.\footnote{\url{https://huggingface.co/OpenGVLab/InternVL3_5-2B}; \url{https://huggingface.co/OpenGVLab/InternVL3_5-8B}; \url{https://huggingface.co/Qwen/Qwen3.5-9B}.} Gemini~2.5 Flash and Pro were accessed through Google's API under its terms of service. The explanatory illustrations in Figures~\ref{fig:rulesmith_overview} and~\ref{fig:method} were generated with Nano Banana Pro (Google) and ChatGPT-5.2 (OpenAI), respectively, under the providers' terms; they convey the pipeline only, and no data figure or reported number was produced with these tools. Game-rollout figures are rendered by the released visualization code.

%% file: tables/all_reevaluations.tex
\begin{table}[htbp]
\papertablestyle
\centering\caption{\textbf{Full 200-/300-game re-evaluations.} $k_E$ is the number of Empire wins from the evaluation logs; none of these batches contains a draw. Losses use counts rather than rounded percentages. Each configuration's two batches use separate rollout seeds.}
\label{tab:all_reevaluations}
\begin{tabular}{lrrrrrrr}
\toprule
Search / run & $n$ & $k_E$ & E (\%) & N (\%) & $\widehat{\mathcal L}$ & E 95\% CI & $p$ \\
\midrule
GP run 1 & 200 & 103 & 51.5 & 48.5 & 0.030 & [44.6, 58.3] & 0.72 \\
GP run 1 & 300 & 155 & 51.7 & 48.3 & 0.033 & [46.0, 57.3] & 0.60 \\
GP run 2 & 200 & 98 & 49.0 & 51.0 & 0.020 & [42.2, 55.9] & 0.83 \\
GP run 2 & 300 & 148 & 49.3 & 50.7 & 0.013 & [43.7, 55.0] & 0.86 \\
GP run 3 & 200 & 103 & 51.5 & 48.5 & 0.030 & [44.6, 58.3] & 0.72 \\
GP run 3 & 300 & 158 & 52.7 & 47.3 & 0.053 & [47.0, 58.2] & 0.39 \\
NTBEA & 200 & 98 & 49.0 & 51.0 & 0.020 & [42.2, 55.9] & 0.83 \\
NTBEA & 300 & 144 & 48.0 & 52.0 & 0.040 & [42.4, 53.6] & 0.53 \\
TPE & 200 & 103 & 51.5 & 48.5 & 0.030 & [44.6, 58.3] & 0.72 \\
TPE & 300 & 163 & 54.3 & 45.7 & 0.087 & [48.7, 59.9] & 0.15 \\
GP-LCB & 200 & 115 & 57.5 & 42.5 & 0.150 & [50.6, 64.1] & 0.04 \\
GP-LCB & 300 & 173 & 57.7 & 42.3 & 0.153 & [52.0, 63.1] & 0.009 \\
\bottomrule
\end{tabular}
\end{table}

%% file: tables/original_intervals.tex
\begin{table}[htbp]
\papertablestyle
\centering\caption{Wilson 95\% intervals for every distinct win rate in the 100-game evaluations of Tables~\ref{tab:ewin_nwin_matrix}, \ref{tab:cross_llm}, \ref{tab:ablation_optimization}, \ref{tab:supp_vs_random}, \ref{tab:supp_vs_rule}, \ref{tab:supp_human_vs_opt}, \ref{tab:supp_system12}, \ref{tab:supp_asymmetric}, and~\ref{tab:extended}, under the 100-game protocol of Section~\ref{subsec:implementation}. A rate of $r$\% corresponds to $r$ wins. For no-draw two-faction outcomes, the Nomads interval is $[100-u,100-l]$ when the Empire interval is $[l,u]$; for three factions, each faction's rate is listed.}
\label{tab:original_intervals}
\begin{tabular}{rr@{\hspace{14pt}}rr@{\hspace{14pt}}rr}
\toprule
Rate (\%) & 95\% CI (\%) & Rate (\%) & 95\% CI (\%) & Rate (\%) & 95\% CI (\%) \\
\midrule
0 & [0.0, 3.7] & 6 & [2.8, 12.5] & 13 & [7.8, 21.0] \\
22 & [15.0, 31.1] & 24 & [16.7, 33.2] & 26 & [18.4, 35.4] \\
29 & [21.0, 38.5] & 31 & [22.8, 40.6] & 32 & [23.7, 41.7] \\
33 & [24.6, 42.7] & 34 & [25.5, 43.7] & 35 & [26.4, 44.7] \\
37 & [28.2, 46.8] & 38 & [29.1, 47.8] & 41 & [31.9, 50.8] \\
45 & [35.6, 54.8] & 46 & [36.6, 55.7] & 47 & [37.5, 56.7] \\
48 & [38.5, 57.7] & 49 & [39.4, 58.7] & 50 & [40.4, 59.6] \\
51 & [41.3, 60.6] & 52 & [42.3, 61.5] & 53 & [43.3, 62.5] \\
55 & [45.2, 64.4] & 61 & [51.2, 70.0] & 65 & [55.3, 73.6] \\
68 & [58.3, 76.3] & 73 & [63.6, 80.7] & 75 & [65.7, 82.5] \\
81 & [72.2, 87.5] & 91 & [83.8, 95.2] & 100 & [96.3, 100.0] \\
\bottomrule
\end{tabular}
\end{table}

%% file: appendix_prompts.tex
\subsection{Complete Rulebook}
The rulebook has ten entries, reproduced verbatim (with whitespace normalized). Entries name rule parameters (for example, \texttt{empire\_unit\_production\_cost}) but never state their values. Two entries are inconsistent with the engine: Rule~1 gives a typical length of 10 turns, whereas all main experiments use a 16-turn cap that is shown in the prompt, and Rule~2 states that Cavalry can \texttt{GATHER}, whereas Rule~6 and the engine do not allow it. The legal-action lists in the prompt follow the engine.
\begin{lstlisting}
Rule 1 - Game Setup: CivMini is played on a 7x7 grid map between two civilizations: Empire and Nomads. Empire starts
with one city at position (1,1), Nomads starts with one city at position (5,5). Both cities are immobile units that
can be attacked. The game lasts for a maximum number of turns (typically 10). Players take turns alternating.
IMPORTANT: Only ONE unit can occupy a cell at any time - no two units (friendly or enemy) can share the same cell.

Rule 2 - Unit Types: Empire has THREE unit types: (1) City - immobile, can produce resources or units, can be
attacked; (2) Farmer - can only GATHER resources and MOVE, cannot battle; (3) Soldier - can only BATTLE and MOVE,
cannot gather. Nomads has TWO unit types: (1) City - same as Empire city; (2) Cavalry - versatile units that can
GATHER, BATTLE, and MOVE. All units (including cities) have hit points and can be destroyed.

Rule 3 - Map: The 7x7 map is a simple grid with no terrain or resource types. All cells are identical. IMPORTANT: Each
cell can contain AT MOST ONE unit (no stacking). Units cannot move to occupied cells.

Rule 4 - Turn Structure: Each turn consists of: (1) Player action phase where EACH UNIT (including cities) of the
current player performs ONE action (multi-action mode), (2) Switch to next player. After both players have acted, the
turn counter increments and all units reset their movement points. In single-action mode, only one unit acts per turn.

Rule 5 - City Actions: Cities are special immobile units that can perform two types of actions: (1) PRODUCE_RESOURCE -
generate resources (empire_city_resource_production for Empire, nomads_city_resource_production for Nomads), (2)
PRODUCE_UNIT - spend resources to create a new unit. Unit costs differ by civilization (empire_unit_production_cost
and nomads_unit_production_cost). Empire cities can produce Farmers (gather only) or Soldiers (combat only). Nomads
cities can only produce Cavalry (both gather and combat). New units appear in an adjacent empty cell.

Rule 6 - Resource Economy: Empire gains resources through GATHER action (Farmers only). Farmers have
empire_farmer_gather_amount gathering rate. Nomads CANNOT gather - instead, Nomads gain resources by KILLING enemy
units. When Nomads destroy an enemy, they receive nomads_kill_resource_gain resources. This creates asymmetric
economies: Empire is peaceful gatherers, Nomads are aggressive raiders.

Rule 7 - Movement Rules: MOVE action allows a unit to move up to its move_points in Manhattan distance (up, down,
left, right - no diagonals). Empire units (Farmers and Soldiers) have empire_unit_move_points (typically 1). Cavalry
have nomads_cavalry_move_points (typically 2, meaning they can move UP TO 2 cells per turn for greater mobility).
Cities have 0 movement points (immobile). CRITICAL: A cell can only contain ONE unit - moves are blocked if ANY unit
(friendly or enemy) is at the destination. Units cannot stack.

Rule 8 - Combat System: BATTLE action attacks an enemy unit in an ADJACENT cell (range 1, up/down/left/right). Only
Soldiers, Cavalry, and Cities can battle. Farmers cannot battle. Damage calculation: base_damage
(empire_battle_base_damage for Empire, nomads_battle_base_damage for Nomads) multiplied by battle_bonus (unified for
both sides). When a unit's HP reaches 0, it is destroyed and removed from the map. Cities have the same HP as their
civilization's combat units (empire_soldier_hp or nomads_cavalry_hp). The attacker gains a battle victory count.

Rule 9 - Victory Conditions: The game ends immediately if one player's CITY is destroyed - the other player wins
instantly. If both cities survive to max_turns, the winner is determined by final score. Final score = resources *
score_per_resource + battles_won * score_per_battle_won + surviving_units * score_per_surviving_unit. Cities are not
scored separately since each side has exactly one.

Rule 10 - Strategic Considerations: Empire has peaceful economy (Farmers gather resources) but must protect Farmers
with Soldiers. Nomads have aggressive economy (gain resources ONLY by killing enemies) forcing them into combat from
the start. Cavalry can move 2 cells per turn, allowing rapid strikes and retreats. Empire can choose defensive play,
but Nomads MUST attack to gain resources for army expansion. This asymmetry creates distinct playstyles: Empire builds
economy then army, Nomads use mobility to raid immediately. Destroying enemy city is instant victory. Cities produce
resources OR units each turn.

\end{lstlisting}
\subsection{System Prompts}
Each faction receives a fixed system prompt; the Empire version is shown below, and the Nomads version differs only in the faction name and the subsequent trait block.
\begin{lstlisting}
You are playing CivMini as the Empire civilization.

Your goal is to WIN the game by DEFEATING your opponent!

CRITICAL: Try your best to attack the units and city of the opponent to win the game! 
The enemy city is your PRIMARY TARGET - destroying it means INSTANT VICTORY!

Strategy:
- ATTACK enemy units and cities aggressively
- Gather resources to build an army
- Expand your forces to overwhelm the enemy
- Protect your own city from enemy attacks

You will be given the current game state and legal actions.
Choose actions based on your civilization's strengths:

Empire Traits:
- STRONGER in combat (battle bonus)
- Economy: GATHER resources with farmers
- Your city at (1,1) TOP-LEFT. Enemies come from (5,5) BOTTOM-RIGHT.

ACTIONS:
- SOLDIERS: MOVE right/down (X+1 or Y+1) to intercept enemies
- FARMERS: GATHER resources when safe
- If enemy adjacent: BATTLE immediately!

Respond with a valid JSON action object. Prioritize military aggression!
\end{lstlisting}
Nomads trait block:
\begin{lstlisting}
Nomads Traits:
- FAST cavalry (mobile raiders) - CAN MOVE 2 CELLS PER TURN!
- Economy: KILL enemies for resources
- Your city at (5,5) BOTTOM-RIGHT. Empire city at (1,1) TOP-LEFT.

STRATEGY:
- PRIMARY GOAL: Destroy Empire city at (1,1)!
- CAVALRY: Move towards Empire CITY (not just left!)
- Use your SPEED ADVANTAGE - Empire units only move 1 cell!
- If enemy adjacent: BATTLE immediately!
- DEFEND: If Empire soldiers approach (5,5), intercept them!
- Target priority: City > Farmers > Soldiers
\end{lstlisting}
\subsection{Example Decision Prompt}
The user prompt below is rendered by the released code for the Empire's first decision under the configuration in Table~\ref{tab:supp_human_vs_opt}; emojis in the strategy-guide banner are replaced by \texttt{[emoji]}. The ``Cities: 0'' line is a display defect in the released code (the city is listed as a unit). The block after ``COPY THESE EXACT ACTIONS'' is produced by the heuristic strategy guide from unit positions alone.
\begin{lstlisting}
RELEVANT RULES:
1. Rule 4 - Turn Structure: Each turn consists of: (1) Player action phase where EACH UNIT 
    (including cities) of the current player performs ONE action (multi-action mode), (2) Switch 
    to next player. After both players have acted, the turn counter increments and all units 
    reset their movement points. In single-action mode, only one unit acts per turn.

2. Rule 5 - City Actions: Cities are special immobile units that can perform two types of 
    actions: (1) PRODUCE_RESOURCE - generate resources (empire_city_resource_production for Empire, 
    nomads_city_resource_production for Nomads), (2) PRODUCE_UNIT - spend resources to create a 
    new unit. Unit costs differ by civilization (empire_unit_production_cost and nomads_unit_production_cost). 
    Empire cities can produce Farmers (gather only) or Soldiers (combat only). Nomads cities can 
    only produce Cavalry (both gather and combat). New units appear in an adjacent empty cell.

3. Rule 10 - Strategic Considerations: Empire has peaceful economy (Farmers gather resources) 
    but must protect Farmers with Soldiers. Nomads have aggressive economy (gain resources ONLY 
    by killing enemies) forcing them into combat from the start. Cavalry can move 2 cells per turn, 
    allowing rapid strikes and retreats. Empire can choose defensive play, but Nomads MUST attack 
    to gain resources for army expansion. This asymmetry creates distinct playstyles: Empire builds 
    economy then army, Nomads use mobility to raid immediately. Destroying enemy city is instant 
    victory. Cities produce resources OR units each turn.

CURRENT GAME STATE:
Turn: 0/16
Your Civilization: Empire
Resources: 10.0
Cities: 0
Battles Won: 0

ENEMY (Nomads) UNITS:
  nomads_city: (5,5) Type=city HP=12.0
  nomads_cavalry_0: (4,5) Type=cavalry HP=12.0
  nomads_cavalry_1: (5,4) Type=cavalry HP=12.0

==================================================
STRATEGY GUIDE (FOLLOW THIS!):
==================================================
[emoji] EMPIRE STRATEGY [emoji]

YOUR CITY: (1,1) - PROTECT IT!
ENEMY CITY: (5,5) - ATTACK TARGET!

[emoji] No immediate threats - push towards enemy city!

COPY THESE EXACT ACTIONS:

"empire_city": {"action_type": "PRODUCE_UNIT", "produce_unit_type": "soldier", "to": {"x": 1, "y": 2}}
  ^ Produce soldier for defense!
"empire_farmer_0": {"action_type": "GATHER"}
  ^ Gather resources safely
"empire_soldier_0": {"action_type": "MOVE", "to": {"x": 1, "y": 0}}
  ^ ATTACK: Move towards Nomads city! (9 cells away)

REMEMBER: COPY THE EXACT JSON ABOVE!
GOAL: Protect city at (1,1) AND attack enemy city at (5,5)!
==================================================

YOUR UNITS (each needs an action):

empire_city:
  Position: (1, 1)
  HP: 9.0, Move Points: 0.0
  Legal actions: PRODUCE_UNIT, PRODUCE_RESOURCE
  Available production positions: (1,2), (1,0), (0,1)

empire_farmer_0:
  Position: (2, 1)
  HP: 5.0, Move Points: 1.0
  Legal actions: MOVE, GATHER

empire_soldier_0:
  Position: (0, 0)
  HP: 9.0, Move Points: 1.0
  Legal actions: MOVE

OUTPUT FORMAT:
Return a JSON object with actions for ALL units:
{
  "unit_id_1": {"action_type": "GATHER"},
  "unit_id_2": {"action_type": "MOVE", "to": {"x": 1, "y": 1}},
  "unit_id_3": {"action_type": "BATTLE", "to": {"x": 2, "y": 2}, "target_civ": "Nomads"},
  "empire_city": {"action_type": "PRODUCE_UNIT", "produce_unit_type": "soldier", "to": {"x": 1, "y": 2}}
}

NOTE: For PRODUCE_UNIT, specify the target position with 'to': {'x': X, 'y': Y}
      The position must be adjacent to the city and empty!

YOUR ACTIONS:
\end{lstlisting}

%% file: checklist.tex
\clearpage
\section*{NeurIPS Paper Checklist}
\begin{enumerate}

\item {\bf Claims}
    \item[] Question: Do the main claims made in the abstract and introduction accurately reflect the paper's contributions and scope?
    \item[] Answer: \answerYes{}
    \item[] Justification: The abstract and introduction state the framework, the CivMini testbed, and the empirical scope. Claims are limited to the tested agent populations and sample sizes.
    \item[] Guidelines:
    \begin{itemize}
        \item The answer [N/A] means that the abstract and introduction do not include the claims made in the paper.
        \item The abstract and/or introduction should clearly state the claims made, including the contributions made in the paper and important assumptions and limitations. A [No] or [N/A] answer to this question will not be perceived well by the reviewers.
        \item The claims made should match theoretical and experimental results, and reflect how much the results can be expected to generalize to other settings.
        \item It is fine to include aspirational goals as motivation as long as it is clear that these goals are not attained by the paper.
    \end{itemize}

\item {\bf Limitations}
    \item[] Question: Does the paper discuss the limitations of the work performed by the authors?
    \item[] Answer: \answerYes{}
    \item[] Justification: Section~\ref{sec:conclusion} and Appendix~\ref{sec:supp_limitations} discuss population dependence of balance, the role of the prompt's strategy guide in the evaluated population, statistical scope, designer-specified rule spaces, the simplified game, and the absence of optimality guarantees.
    \item[] Guidelines:
    \begin{itemize}
        \item The answer [N/A] means that the paper has no limitation while the answer [No] means that the paper has limitations, but those are not discussed in the paper.
        \item The authors are encouraged to create a separate ``Limitations'' section in their paper.
        \item The paper should point out any strong assumptions and how robust the results are to violations of these assumptions (e.g., independence assumptions, noiseless settings, model well-specification, asymptotic approximations only holding locally). The authors should reflect on how these assumptions might be violated in practice and what the implications would be.
        \item The authors should reflect on the scope of the claims made, e.g., if the approach was only tested on a few datasets or with a few runs. In general, empirical results often depend on implicit assumptions, which should be articulated.
        \item The authors should reflect on the factors that influence the performance of the approach. For example, a facial recognition algorithm may perform poorly when image resolution is low or images are taken in low lighting. Or a speech-to-text system might not be used reliably to provide closed captions for online lectures because it fails to handle technical jargon.
        \item The authors should discuss the computational efficiency of the proposed algorithms and how they scale with dataset size.
        \item If applicable, the authors should discuss possible limitations of their approach to address problems of privacy and fairness.
        \item While the authors might fear that complete honesty about limitations might be used by reviewers as grounds for rejection, a worse outcome might be that reviewers discover limitations that aren't acknowledged in the paper. The authors should use their best judgment and recognize that individual actions in favor of transparency play an important role in developing norms that preserve the integrity of the community. Reviewers will be specifically instructed to not penalize honesty concerning limitations.
    \end{itemize}

\item {\bf Theory assumptions and proofs}
    \item[] Question: For each theoretical result, does the paper provide the full set of assumptions and a complete (and correct) proof?
    \item[] Answer: \answerYes{}
    \item[] Justification: Appendix~\ref{sec:supp_theory} states all assumptions and gives proofs for Proposition~\ref{prop:symmetry}, Lemma~\ref{lemma:concentration_of_loss}, and Proposition~\ref{prop:idealized_gp_lcb}; the remarks state that the regret bound covers an idealized LCB rule, not the implemented acquisition or adaptive replication.
    \item[] Guidelines:
    \begin{itemize}
        \item The answer [N/A] means that the paper does not include theoretical results.
        \item All the theorems, formulas, and proofs in the paper should be numbered and crossreferenced.
        \item All assumptions should be clearly stated or referenced in the statement of any theorems.
        \item The proofs can either appear in the main paper or the supplemental material, but if they appear in the supplemental material, the authors are encouraged to provide a short proof sketch to provide intuition.
        \item Inversely, any informal proof provided in the core of the paper should be complemented by formal proofs provided in appendix or supplemental material.
        \item Theorems and Lemmas that the proof relies upon should be properly referenced.
    \end{itemize}

\item {\bf Experimental result reproducibility}
    \item[] Question: Does the paper fully disclose all the information needed to reproduce the main experimental results of the paper to the extent that it affects the main claims and/or conclusions of the paper (regardless of whether the code and data are provided or not)?
    \item[] Answer: \answerYes{}
    \item[] Justification: Section~\ref{subsec:implementation}, Table~\ref{tab:parameters}, Table~\ref{tab:settings}, and Appendix~\ref{sec:supp_reproducibility} give the models, decoding, prompts and rulebook verbatim, optimizer settings, seeds, budgets, and evaluation protocol.
    \item[] Guidelines:
    \begin{itemize}
        \item The answer [N/A] means that the paper does not include experiments.
        \item If the paper includes experiments, a [No] answer to this question will not be perceived well by the reviewers: Making the paper reproducible is important, regardless of whether the code and data are provided or not.
        \item If the contribution is a dataset and/or model, the authors should describe the steps taken to make their results reproducible or verifiable.
        \item Depending on the contribution, reproducibility can be accomplished in various ways. For example, if the contribution is a novel architecture, describing the architecture fully might suffice, or if the contribution is a specific model and empirical evaluation, it may be necessary to either make it possible for others to replicate the model with the same dataset, or provide access to the model. In general. releasing code and data is often one good way to accomplish this, but reproducibility can also be provided via detailed instructions for how to replicate the results, access to a hosted model (e.g., in the case of a large language model), releasing of a model checkpoint, or other means that are appropriate to the research performed.
        \item While NeurIPS does not require releasing code, the conference does require all submissions to provide some reasonable avenue for reproducibility, which may depend on the nature of the contribution. For example (a) If the contribution is primarily a new algorithm, the paper should make it clear how to reproduce that algorithm. (b) If the contribution is primarily a new model architecture, the paper should describe the architecture clearly and fully. (c) If the contribution is a new model (e.g., a large language model), then there should either be a way to access this model for reproducing the results or a way to reproduce the model (e.g., with an open-source dataset or instructions for how to construct the dataset). (d) We recognize that reproducibility may be tricky in some cases, in which case authors are welcome to describe the particular way they provide for reproducibility. In the case of closed-source models, it may be that access to the model is limited in some way (e.g., to registered users), but it should be possible for other researchers to have some path to reproducing or verifying the results.
    \end{itemize}

\item {\bf Open access to data and code}
    \item[] Question: Does the paper provide open access to the data and code, with sufficient instructions to faithfully reproduce the main experimental results, as described in supplemental material?
    \item[] Answer: \answerYes{}
    \item[] Justification: Code for the main experiments, the prompts, and the analysis scripts are released under the MIT license at \url{https://github.com/Adonis-galaxy/RuleSmith}, with instructions to run the optimization and the held-out evaluations.
    \item[] Guidelines:
    \begin{itemize}
        \item The answer [N/A] means that paper does not include experiments requiring code.
        \item Please see the NeurIPS code and data submission guidelines (https://neurips.cc/ public/guides/CodeSubmissionPolicy) for more details.
        \item While we encourage the release of code and data, we understand that this might not be possible, so [No] is an acceptable answer. Papers cannot be rejected simply for not including code, unless this is central to the contribution (e.g., for a new open-source benchmark).
        \item The instructions should contain the exact command and environment needed to run to reproduce the results. See the NeurIPS code and data submission guidelines (https: //neurips.cc/public/guides/CodeSubmissionPolicy) for more details.
        \item The authors should provide instructions on data access and preparation, including how to access the raw data, preprocessed data, intermediate data, and generated data, etc.
        \item The authors should provide scripts to reproduce all experimental results for the new proposed method and baselines. If only a subset of experiments are reproducible, they should state which ones are omitted from the script and why.
        \item At submission time, to preserve anonymity, the authors should release anonymized versions (if applicable).
        \item Providing as much information as possible in supplemental material (appended to the paper) is recommended, but including URLs to data and code is permitted.
    \end{itemize}

\item {\bf Experimental setting/details}
    \item[] Question: Does the paper specify all the training and test details (e.g., data splits, hyperparameters, how they were chosen, type of optimizer) necessary to understand the results?
    \item[] Answer: \answerYes{}
    \item[] Justification: Section~\ref{subsec:implementation} and Table~\ref{tab:settings} specify the optimizer, acquisition, budgets, adaptive-sampling bounds, discretization, and the held-out evaluation protocol, and state which settings were chosen by convention rather than tuned.
    \item[] Guidelines:
    \begin{itemize}
        \item The answer [N/A] means that the paper does not include experiments.
        \item The experimental setting should be presented in the core of the paper to a level of detail that is necessary to appreciate the results and make sense of them.
        \item The full details can be provided either with the code, in appendix, or as supplemental material.
    \end{itemize}

\item {\bf Experiment statistical significance}
    \item[] Question: Does the paper report error bars suitably and correctly defined or other appropriate information about the statistical significance of the experiments?
    \item[] Answer: \answerYes{}
    \item[] Justification: Win rates of the additional evaluations carry Wilson 95\% intervals and exact two-sided binomial tests; repeatability is assessed with three independent optimization runs, and the 100-game original results are given Wilson intervals in Table~\ref{tab:original_intervals}. Definitions and assumptions are in Appendix~\ref{sec:supp_statistics}. Original ablations remain single runs, as stated.
    \item[] Guidelines:
    \begin{itemize}
        \item The answer [N/A] means that the paper does not include experiments.
        \item The authors should answer [Yes] if the results are accompanied by error bars, confidence intervals, or statistical significance tests, at least for the experiments that support the main claims of the paper.
        \item The factors of variability that the error bars are capturing should be clearly stated (for example, train/test split, initialization, random drawing of some parameter, or overall run with given experimental conditions).
        \item The method for calculating the error bars should be explained (closed form formula, call to a library function, bootstrap, etc.)
        \item The assumptions made should be given (e.g., Normally distributed errors).
        \item It should be clear whether the error bar is the standard deviation or the standard error of the mean.
        \item It is OK to report 1-sigma error bars, but one should state it. The authors should preferably report a 2-sigma error bar than state that they have a 96\% CI, if the hypothesis of Normality of errors is not verified.
        \item For asymmetric distributions, the authors should be careful not to show in tables or figures symmetric error bars that would yield results that are out of range (e.g., negative error rates).
        \item If error bars are reported in tables or plots, the authors should explain in the text how they were calculated and reference the corresponding figures or tables in the text.
    \end{itemize}

\item {\bf Experiments compute resources}
    \item[] Question: For each experiment, does the paper provide sufficient information on the computer resources (type of compute workers, memory, time of execution) needed to reproduce the experiments?
    \item[] Answer: \answerYes{}
    \item[] Justification: Section~\ref{subsec:implementation} states the hardware of the original experiments (8 NVIDIA A100 80GB) and of the added runs (NVIDIA RTX 6000 Ada), and the wall-clock time of a 100-iteration optimization.
    \item[] Guidelines:
    \begin{itemize}
        \item The answer [N/A] means that the paper does not include experiments.
        \item The paper should indicate the type of compute workers CPU or GPU, internal cluster, or cloud provider, including relevant memory and storage.
        \item The paper should provide the amount of compute required for each of the individual experimental runs as well as estimate the total compute.
        \item The paper should disclose whether the full research project required more compute than the experiments reported in the paper (e.g., preliminary or failed experiments that didn't make it into the paper).
    \end{itemize}

\item {\bf Code of ethics}
    \item[] Question: Does the research conducted in the paper conform, in every respect, with the NeurIPS Code of Ethics https://neurips.cc/public/EthicsGuidelines?
    \item[] Answer: \answerYes{}
    \item[] Justification: The work uses a synthetic game and small gameplay studies in which the authors themselves were the players; only game outcomes were recorded, and no personal or sensitive data were collected (Appendix~\ref{sec:supp_user_study}).
    \item[] Guidelines:
    \begin{itemize}
        \item The answer [N/A] means that the authors have not reviewed the NeurIPS Code of Ethics.
        \item If the authors answer [No], they should explain the special circumstances that require a deviation from the Code of Ethics.
        \item The authors should make sure to preserve anonymity (e.g., if there is a special consideration due to laws or regulations in their jurisdiction).
    \end{itemize}

\item {\bf Broader impacts}
    \item[] Question: Does the paper discuss both potential positive societal impacts and negative societal impacts of the work performed?
    \item[] Answer: \answerYes{}
    \item[] Justification: Appendix~\ref{sec:supp_impact} discusses positive uses (design tools, analysis of rule-driven systems) and risks (over-optimizing rules to the disadvantage of participants), with mitigations.
    \item[] Guidelines:
    \begin{itemize}
        \item The answer [N/A] means that there is no societal impact of the work performed.
        \item If the authors answer [N/A] or [No], they should explain why their work has no societal impact or why the paper does not address societal impact.
        \item Examples of negative societal impacts include potential malicious or unintended uses (e.g., disinformation, generating fake profiles, surveillance), fairness considerations (e.g., deployment of technologies that could make decisions that unfairly impact specific groups), privacy considerations, and security considerations.
        \item The conference expects that many papers will be foundational research and not tied to particular applications, let alone deployments. However, if there is a direct path to any negative applications, the authors should point it out. For example, it is legitimate to point out that an improvement in the quality of generative models could be used to generate Deepfakes for disinformation. On the other hand, it is not needed to point out that a generic algorithm for optimizing neural networks could enable people to train models that generate Deepfakes faster.
        \item The authors should consider possible harms that could arise when the technology is being used as intended and functioning correctly, harms that could arise when the technology is being used as intended but gives incorrect results, and harms following from (intentional or unintentional) misuse of the technology.
        \item If there are negative societal impacts, the authors could also discuss possible mitigation strategies (e.g., gated release of models, providing defenses in addition to attacks, mechanisms for monitoring misuse, mechanisms to monitor how a system learns from feedback over time, improving the efficiency and accessibility of ML).
    \end{itemize}

\item {\bf Safeguards}
    \item[] Question: Does the paper describe safeguards that have been put in place for responsible release of data or models that have a high risk for misuse (e.g., pre-trained language models, image generators, or scraped datasets)?
    \item[] Answer: \answerNA{}
    \item[] Justification: The paper releases a game environment and an optimization pipeline, not pretrained models or scraped data with high misuse risk.
    \item[] Guidelines:
    \begin{itemize}
        \item The answer [N/A] means that the paper poses no such risks.
        \item Released models that have a high risk for misuse or dual-use should be released with necessary safeguards to allow for controlled use of the model, for example by requiring that users adhere to usage guidelines or restrictions to access the model or implementing safety filters.
        \item Datasets that have been scraped from the Internet could pose safety risks. The authors should describe how they avoided releasing unsafe images.
        \item We recognize that providing effective safeguards is challenging, and many papers do not require this, but we encourage authors to take this into account and make a best faith effort.
    \end{itemize}

\item {\bf Licenses for existing assets}
    \item[] Question: Are the creators or original owners of assets (e.g., code, data, models), used in the paper, properly credited and are the license and terms of use explicitly mentioned and properly respected?
    \item[] Answer: \answerYes{}
    \item[] Justification: InternVL3.5 is cited, and InternVL3.5 and Qwen3.5 are released under Apache-2.0 (model cards in Appendix~\ref{sec:supp_reproducibility}); Gemini~2.5 was accessed under Google's API terms; scikit-optimize (BSD-3-Clause) and Optuna (MIT) are cited; the figure-generation tools are credited (Appendix~\ref{sec:supp_reproducibility}).
    \item[] Guidelines:
    \begin{itemize}
        \item The answer [N/A] means that the paper does not use existing assets.
        \item The authors should cite the original paper that produced the code package or dataset.
        \item The authors should state which version of the asset is used and, if possible, include a URL.
        \item The name of the license (e.g., CC-BY 4.0) should be included for each asset.
        \item For scraped data from a particular source (e.g., website), the copyright and terms of service of that source should be provided.
        \item If assets are released, the license, copyright information, and terms of use in the package should be provided. For popular datasets, paperswithcode.com/datasets has curated licenses for some datasets. Their licensing guide can help determine the link of a dataset.
        \item For existing datasets that are re-packaged, both the original license and the license of the derived asset (if it has changed) should be provided.
        \item If this information is not available online, the authors are encouraged to reach out to the asset's creators.
    \end{itemize}

\item {\bf New assets}
    \item[] Question: Are new assets introduced in the paper well documented and is the documentation provided alongside the assets?
    \item[] Answer: \answerYes{}
    \item[] Justification: CivMini and the RuleSmith pipeline are documented in Section~\ref{sec:method} and Appendix~\ref{sec:supp_reproducibility} (complete rulebook, prompts, and settings) and in the repository README, under the MIT license.
    \item[] Guidelines:
    \begin{itemize}
        \item The answer [N/A] means that the paper does not release new assets.
        \item Researchers should communicate the details of the dataset/code/model as part of their submissions via structured templates. This includes details about training, license, limitations, etc.
        \item The paper should discuss whether and how consent was obtained from people whose asset is used.
        \item At submission time, remember to anonymize your assets (if applicable). You can either create an anonymized URL or include an anonymized zip file.
    \end{itemize}

\item {\bf Crowdsourcing and research with human subjects}
    \item[] Question: For crowdsourcing experiments and research with human subjects, does the paper include the full text of instructions given to participants and screenshots, if applicable, as well as details about compensation (if any)?
    \item[] Answer: \answerYes{}
    \item[] Justification: Appendix~\ref{sec:supp_user_study} describes both human studies: the participants were authors of this paper, their only instructions were the natural-language rulebook reproduced in Appendix~\ref{sec:supp_reproducibility}, they played through a text interface showing the game state, and they received no compensation.
    \item[] Guidelines:
    \begin{itemize}
        \item The answer [N/A] means that the paper does not involve crowdsourcing nor research with human subjects.
        \item Including this information in the supplemental material is fine, but if the main contribution of the paper involves human subjects, then as much detail as possible should be included in the main paper.
        \item According to the NeurIPS Code of Ethics, workers involved in data collection, curation, or other labor should be paid at least the minimum wage in the country of the data collector.
    \end{itemize}

\item {\bf Institutional review board (IRB) approvals or equivalent for research with human subjects}
    \item[] Question: Does the paper describe potential risks incurred by study participants, whether such risks were disclosed to the subjects, and whether Institutional Review Board (IRB) approvals (or an equivalent approval/review based on the requirements of your country or institution) were obtained?
    \item[] Answer: \answerYes{}
    \item[] Justification: Appendix~\ref{sec:supp_user_study} states that all participants were authors of this paper, that the sessions posed no risk beyond ordinary game play and collected no personal data, and that the study was therefore not submitted for institutional review.
    \item[] Guidelines:
    \begin{itemize}
        \item The answer [N/A] means that the paper does not involve crowdsourcing nor research with human subjects.
        \item Depending on the country in which research is conducted, IRB approval (or equivalent) may be required for any human subjects research. If you obtained IRB approval, you should clearly state this in the paper.
        \item We recognize that the procedures for this may vary significantly between institutions and locations, and we expect authors to adhere to the NeurIPS Code of Ethics and the guidelines for their institution.
        \item For initial submissions, do not include any information that would break anonymity (if applicable), such as the institution conducting the review.
    \end{itemize}

\item {\bf Declaration of LLM usage}
    \item[] Question: Does the paper describe the usage of LLMs if it is an important, original, or non-standard component of the core methods in this research? Note that if the LLM is used only for writing, editing, or formatting purposes and does not impact the core methodology, scientific rigor, or originality of the research, declaration is not required.
    \item[] Answer: \answerYes{}
    \item[] Justification: LLMs are the game-playing agents whose self-play defines the balance objective; their use, prompts, and decoding are described in Sections~\ref{subsec:llm_selfplay} and~\ref{subsec:implementation} and Appendix~\ref{sec:supp_reproducibility}.
    \item[] Guidelines:
    \begin{itemize}
        \item The answer [N/A] means that the core method development in this research does not involve LLMs as any important, original, or non-standard components.
        \item Please refer to our LLM policy in the NeurIPS handbook for what should or should not be described.
    \end{itemize}

\end{enumerate}